\documentclass[10pt,twocolumn,letterpaper]{article}

\usepackage{cvpr}              
\definecolor{cvprblue}{rgb}{0.21,0.49,0.74}
\usepackage{multirow}
\usepackage{cuted} 
\usepackage{caption}
\usepackage{makecell}
\usepackage{amsmath}
\usepackage{pifont}
\usepackage{bbding}
\usepackage[table]{xcolor}
\usepackage{booktabs}
\usepackage{colortbl}
\usepackage{array}
\usepackage{nicematrix}
\usepackage[pagebackref,breaklinks,colorlinks,allcolors=cvprblue]{hyperref}
\def\paperID{} 
\def\confName{CVPR}
\def\confYear{2026}
\newcommand*{\affaddr}[1]{#1} 
\newcommand*{\affmark}[1][*]{\textsuperscript{#1}}

\makeatletter
\def\thanks#1{\protected@xdef\@thanks{\@thanks\protect\footnotetext{#1}}}
\makeatother
\title{Interactive Tracking: A Human-in-the-Loop Paradigm with Memory-Augmented Adaptation}

\author{%
Yuqing Huang\affmark[1,2,${\dag}$], Guotian Zeng\affmark[3,1,${\dag}$], Zhenqiao Yuan\affmark[1,2], Zhenyu He\affmark[2,$\ast$], Xin Li\affmark[1,4,$\ast$], \\ Yaowei Wang\affmark[2,1], and Ming-Hsuan Yang\affmark[5,6]\\
\affaddr{\affmark[1]Pengcheng Laboratory}\quad
\affaddr{\affmark[2]Harbin Institute of Technology, Shenzhen}\\
\affaddr{\affmark[3]South China University of Technology}\quad
\affaddr{\affmark[4]Pazhou Lab (Huangpu)}\quad
\affaddr{\affmark[5]UC Merced}
\quad
\affaddr{\affmark[6]Yonsei University}\\
}

\begin{document}
\twocolumn[{%
\renewcommand\twocolumn[1][]{#1}%
\maketitle
\begin{center}
    \centering
    \vspace{-5mm}
    \includegraphics[width=0.9\textwidth]{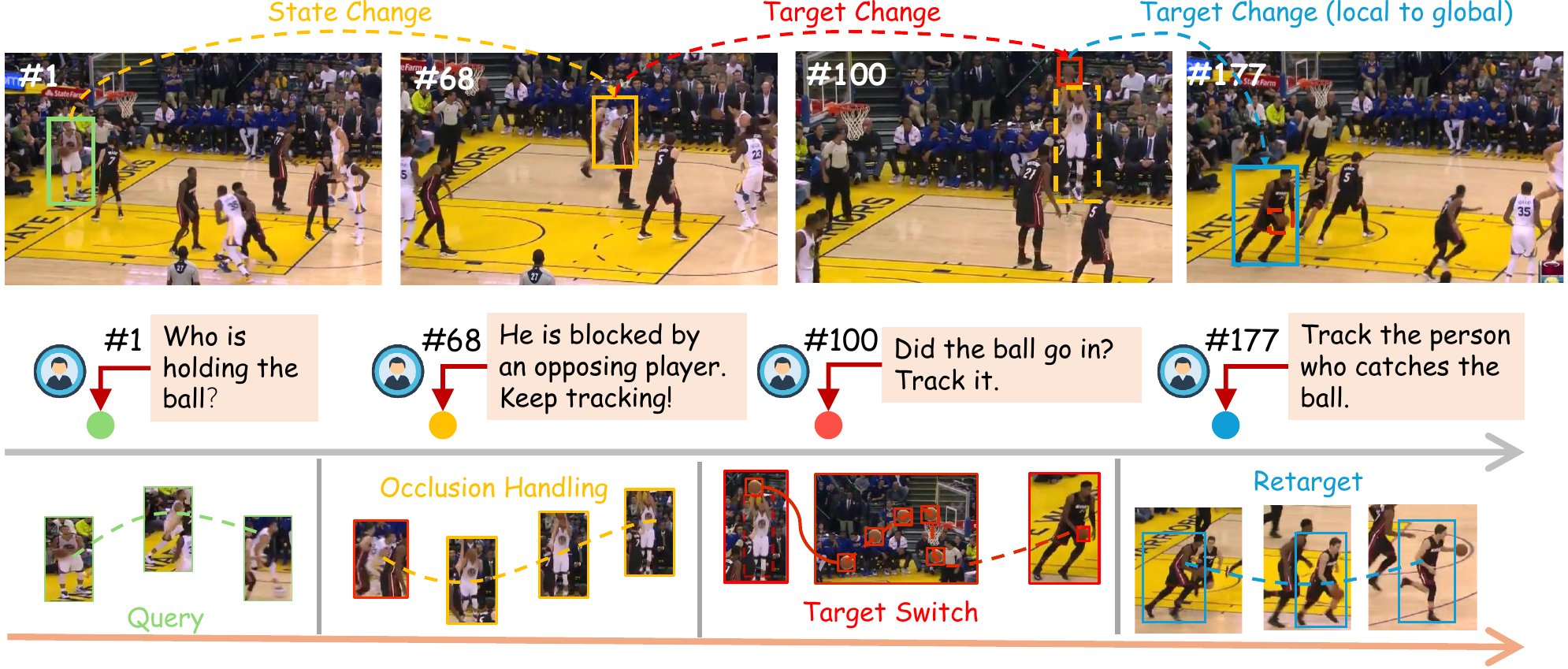} 
    \vspace{-4mm}
    \captionof{figure}{\textbf{Interactive tracking in a basketball sequence.} User prompts guide the tracker through state changes, target changes, and global retargeting, demonstrating the interaction loop required for real-world analysis.}
\label{fig:motivation}
\end{center}%
}]
\begingroup
\renewcommand{\thefootnote}{} 
\footnotetext{$^\dag$ Equal contribution, $\ast$ Corresponding author}
\endgroup


\begin{abstract}
Existing visual trackers mainly operate in a non-interactive, fire-and-forget manner, making them impractical for real-world scenarios that require human-in-the-loop adaptation.
To overcome this limitation, we introduce Interactive Tracking, a new paradigm that allows users to guide the tracker at any time using natural language commands.
To support research in this direction, we make three main contributions.
First, we present InteractTrack, the first large-scale benchmark for interactive tracking, containing 150 videos with dense bounding box annotations and timestamped language instructions.
Second, we propose a comprehensive evaluation protocol and evaluate 25 representative trackers, showing that state-of-the-art methods fail in interactive scenarios—strong performance on conventional benchmarks does not transfer.
Third, we introduce Interactive Memory-Augmented Tracking (IMAT), a new baseline that employs a dynamic memory mechanism to learn from user feedback and update tracking behavior accordingly.
Our benchmark, protocol, and baseline establish a foundation for developing more intelligent, adaptive, and collaborative tracking systems, bridging the gap between automated perception and human guidance.
The full benchmark, tracking results, and analysis are available at \href{https://github.com/NorahGreen/InteractTrack.git}{this URL}.

\end{abstract}

    
    
    
    
    

\section{Introduction}
Visual object tracking (VOT)~\cite{siamfc,ocean} is a cornerstone of computer vision, supporting a wide range of applications including surveillance, autonomous driving, and robotics.
Its objective is to continuously localize a target throughout a video sequence given only its initial state.
However, in real-world scenarios, tracking is rarely a fire-and-forget process—targets of interest often change, and users may wish to guide or correct the tracker during operation.

Fig.~\ref{fig:motivation} illustrates a representative example from a basketball video.
Initially, the tracker follows the player holding the ball; as the play unfolds, the viewer’s attention naturally shifts—from one player to another, then to the fast-moving ball, and finally to a different ball handler.
Such dynamic shifts of focus are natural for human observers but remain unsupported by existing tracking systems, which continue to operate autonomously after initialization.
This disconnect highlights the need for interactive tracking, where human intelligence complements automated perception to handle flexible, real-world situations.

Despite significant progress in autonomous tracking, current paradigms and benchmarks were not designed to support interactive scenarios.
They typically assume a fixed target and lack mechanisms for user intervention or semantic instruction.
Consequently, when the target changes or the tracker drifts, these systems cannot recover without manual reinitialization.
This limitation is especially apparent in complex applications such as sports analysis or UAV monitoring, where context-aware, human-guided tracking is essential for robust and reliable performance.

Unlike conventional tracking, interactive tracking requires the model to continuously respond to user guidance in real time, comprehend natural-language instructions, and dynamically adapt its focus.
This task is substantially more challenging than traditional tracking, as it tightly couples perception, reasoning, and human interaction within a continuous feedback loop.
However, existing tracking paradigms cannot meet these demands.

Conventional Visual Object Tracking (VOT) methods rely solely on appearance cues and lack the ability to interpret or react to user commands.
In contrast, vision–language approaches such as Vision–Language Tracking (VLT) and Referring Video Object Segmentation (RVOS) incorporate natural-language input to specify targets, but they typically perform one-time grounding at initialization or operate offline, preventing them from handling sequential user instructions or supporting real-time, human-in-the-loop interaction.
Moreover, current benchmarks—including VOT~\cite{vot2014}, LaSOT~\cite{lasot}, VideoCube~\cite{hu2022global}, MGIT~\cite{mgit}, and InsT~\cite{trackGPT}—evaluate trackers in purely autonomous settings, offering no means to assess responsiveness or user-guided adaptation.
%
%
These limitations highlight the urgent need for a new tracking paradigm that seamlessly integrates perception, interaction, and reasoning, paving the way toward adaptive and collaborative tracking systems.


To bridge this gap, we introduce a new task—interactive tracking—where a user can dynamically guide a tracker through natural language commands at any point during a video.
To facilitate systematic research in this setting, we present InteractTrack, the first large-scale benchmark explicitly designed for interactive visual object tracking.
The dataset comprises 150 curated videos totaling over 140K frames, spanning six diverse real-world scenarios, including sports analysis, surveillance, UAV tracking, wildlife monitoring, daily activities, and other scenarios.
Each sequence is densely annotated with frame-level bounding boxes and multiple natural-language expressions (approximately 4–5 per video) representing distinct interaction types, such as initialization, drift correction, target switching, and part-level refinement.
By combining temporal continuity with multimodal annotations, InteractTrack enables systematic evaluation of a tracker’s ability to interpret, respond to, and adapt to human guidance under realistic, dynamically evolving conditions.

Alongside the dataset, we establish a comprehensive evaluation protocol specifically tailored to interactive tracking.
Unlike conventional precision and success metrics, our protocol evaluates additional dimensions such as perceptual understanding, responsiveness, and adaptability, which are critical for measuring a tracker’s interactive capability.
We further propose Interactive Memory-Augmented Tracking (IMAT), a new framework that unifies the spatiotemporal consistency of visual tracking with the semantic reasoning of multimodal large language models.
IMAT introduces a dynamic memory mechanism composed of positive and negative banks, enabling the tracker to learn from user feedback by retaining both discriminative cues and distractor features.
This memory-driven adaptation progressively enhances robustness, responsiveness, and semantic alignment during continuous interaction.

Together, InteractTrack and IMAT establish a foundation for developing and evaluating the next generation of interactive visual tracking systems.
The main contributions of this work are:
\begin{itemize}
\item We present InteractTrack, the first large-scale benchmark explicitly designed for interactive visual object tracking, featuring rich multimodal annotations across diverse real-world scenarios.
\item We establish a comprehensive evaluation protocol and perform an extensive analysis of 25 representative trackers, providing strong baselines and revealing the unique challenges of interactive tracking.
\item We propose Interactive Memory-Augmented Tracking (IMAT), a novel framework equipped with a dynamic memory mechanism that learns from user feedback to progressively improve tracking accuracy, robustness, and adaptability.
\end{itemize}

\section{Related Work}

\noindent\textbf{Conventional Tracking.}
Visual Object Tracking (VOT) aims to continuously localize a target initialized by a bounding box in the first frame.
Most modern trackers formulate this task as a feature-matching problem between a static template and a dynamic search region.
Early Siamese-based methods~\cite{siamattn,siamdw,siamrcnn,siammask}, such as SiamFC~\cite{siamfc} and SiamRPN~\cite{siamrpn}, learn appearance similarity to match targets across frames efficiently.
With the introduction of transformers~\cite{transformer}, subsequent trackers~\cite{zeng2024visual,lorat,dreamtrack,aqatrack, swintrack,rtracker, loratv2}, including TransT~\cite{transt}, STARK~\cite{stark}, MixFormer~\cite{mixformer}, ARTrack~\cite{artrack}, and OSTrack~\cite{ostrack}, achieved significant improvements through enhanced global feature interaction and spatiotemporal reasoning.
In parallel, semi-supervised Video Object Segmentation (VOS) extends tracking to the pixel level by providing an initial segmentation mask.
Approaches such as STM~\cite{stm}, AOT~\cite{aot}, XMem~\cite{xmem}, and SAM2~\cite{sam2} propagate the initial mask using memory-based matching, enabling fine-grained object localization.
Despite these advances, both VOT and VOS depend entirely on fixed visual templates defined in the first frame and lack mechanisms for user feedback or semantic adaptation, limiting their applicability in interactive tracking scenarios.

\noindent\textbf{Referring Tracking.}
When explicit visual templates are unavailable, tracking can be guided by natural language descriptions, transforming the problem into a referring-object tracking task.
Depending on the output format, this research area can be divided into Vision–Language Tracking (VLT) and Referring Video Object Segmentation (RVOS).

VLT methods~\cite{li2025dynamic,queryNLT,onevl,allinone,atctrack,memvlt}, such as VLT~\cite{divertmore}, MMTrack~\cite{mmtrack}, JointVLT~\cite{zhou2023joint}, and UVLTrack~\cite{UVLTrack}, employ encoders based on BERT~\cite{bert} or CLIP~\cite{clip} to extract semantic embeddings from text and align them with visual features for cross-modal grounding.
RVOS approaches, including ReferFormer~\cite{referformer}, LBDT~\cite{LBDT}, and R-VOS~\cite{R-VOS}, extend this to the pixel level by incorporating cross-modal transformers or memory mechanisms to enhance spatiotemporal consistency.

Recent studies have generalized referring segmentation and tracking into broader multimodal frameworks.
DETR-like architectures unify segmentation and tracking, while MLLM-based methods such as LISA~\cite{lai2023lisa} and GLaMM~\cite{glamm} introduce reasoning-based segmentation and joint mask–caption generation.
Other models, including VISA~\cite{yan2024visa}, VideoLISA~\cite{videolisa}, and Sa2VA~\cite{sa2va}, explore open-vocabulary and instruction-tuned referring video segmentation, enabling broader visual–language understanding.

Although these approaches have significantly advanced the integration of vision, language, and reasoning, they generally operate offline or rely on single-pass grounding, making them unable to handle sequential user instructions, context updates, or real-time interaction—capabilities essential for human-in-the-loop tracking.

\noindent\textbf{Interactive Tasks.}
Interactive segmentation has evolved from early click-based methods such as FocalClick~\cite{focalclick} and RITM~\cite{ritm} to promptable foundation models that accept diverse user inputs, including points, boxes, and text prompts.
Recent advances such as SAM2~\cite{sam2} enable unified image–video interaction, while LLM-based approaches, including LISA~\cite{lai2023lisa}, GLaMM~\cite{glamm}, and Groundhog~\cite{zhang2024groundhog}, integrate perception and reasoning for text-grounded segmentation.
Further developments, such as PSALM~\cite{zhang2024psalm}, LLM-Seg~\cite{wang2024llm}, and VISA~\cite{yan2024visa}, extend this paradigm toward multimodal and video-level understanding.

Despite these advances, current interactive segmentation~\cite{zhou2023interactive} and grounding systems~\cite{zhuang2025spatial, rex} remain offline and single-pass, lacking the ability to handle sequential user inputs or maintain temporal consistency.
In this work, we explore Interactive Tracking, a new formulation in which user feedback is incorporated at runtime to guide re-localization, target switching, and drift correction.
This real-time, human-in-the-loop paradigm bridges perception, reasoning, and tracking, paving the way for more adaptive and collaborative visual systems.

\begin{table}[tp]
\centering
\caption{\textbf{Comparison of visual tracking benchmarks.} This table summarizes key statistics of each test set and highlights support for interactive evaluation and multi-scenario coverage.}
\label{tab:benchmarks_comparison}
\setlength{\tabcolsep}{3pt} 
\resizebox{\linewidth}{!}{%
\begin{tabular}{lrrrrcc} 
\toprule
\textbf{Benchmark} & \textbf{Year} & \textbf{Videos} & \makecell[c]{\textbf{Mean}\\\textbf{Frames}} &  \makecell[c]{\textbf{Total}\\\textbf{Frames}} &
\makecell[c]{\textbf{Interactive}\\\textbf{Label}} &
\makecell[c]{\textbf{Scenarios}\\\textbf{Label}} \\
\midrule
OTB-2013~\cite{otb2013} & 2013 & 50 & 578 & 29K & \ding{55}  & \ding{55}  \\
OTB-2015~\cite{otb} & 2015 & 100 & 590 & 59K & \ding{55}  & \ding{55}  \\
TC-128~\cite{tc128} & 2015 & 128 & 429 & 55K & \ding{55}  & \ding{55} \\
NUS-PRO~\cite{NUS_PRO} & 2016 & 365 & 371 & 135K & \ding{55}  & \ding{55} \\
UAV123~\cite{uav123}  & 2016 & 123 & 915 & 113K & \ding{55}  & \ding{55} \\
UAV20L~\cite{uav123} & 2016 & 20 & 2,934 & 59K & \ding{55}  & \ding{55} \\
NFS~\cite{nfs} & 2017 & 100 & 3,830 & 383K & \ding{55}  & \ding{55} \\
VOT-2017~\cite{vot2017} & 2017 & 60 & 356 & 21K & \ding{55}  & \ding{55} \\
OxUvA~\cite{oxuav} & 2018 & 366 & 4,235 & 1.55M & \ding{55}  & \ding{55} \\
TrackingNet~\cite{trackingnet} & 2018 & 511 & 441.5 & 225.6K & \ding{55}  & \ding{55} \\
LaSOT~\cite{lasot} & 2019 & 280 & 2,053 & 685.4K & \ding{55}  & \ding{55} \\
GOT-10k~\cite{got10k} & 2021 & 180 & 127 & 22.8K & \ding{55}  & \ding{55} \\
TNL2K~\cite{tnl2k} & 2021 & 700 & 737 & 516K & \ding{55}  & \ding{55} \\
VastTrack~\cite{vasttrack} & 2024 & 3,500 & 106.3 & 372K & \ding{55}  & \ding{55} \\
\midrule
\textbf{InteractTrack} & \textbf{2025} & \textbf{150} & \textbf{947} & \textbf{140K} & \textbf{\checkmark } & \textbf{\checkmark} \\
\bottomrule
\end{tabular}
}
\vspace{-4mm}
\end{table}

\section{InteractTrack Benchmark}

In this section, we present InteractTrack, a benchmark explicitly designed for interactive visual object tracking.
Unlike conventional autonomous tracking settings, InteractTrack captures natural user–tracker interactions, enabling systematic evaluation of how effectively tracking systems interpret, respond to, and adapt to human guidance in dynamic visual environments.

\begin{figure*}[tp]
    \centering
    \includegraphics[width=0.9\textwidth]{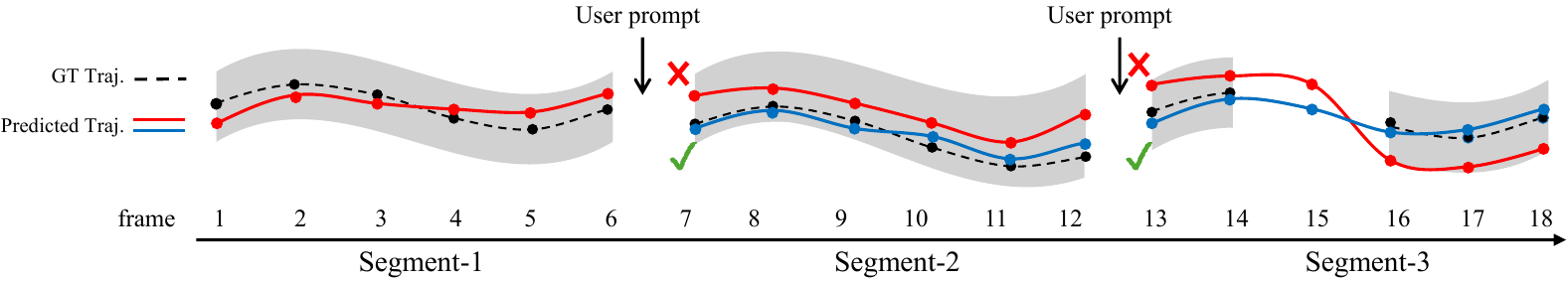}
    \caption{
    \textbf{Overview of the interactive evaluation protocol.}
    Each video is divided into segments by user prompts. 
    Dashed lines denote ground-truth trajectories, and colored curves represent tracker predictions. 
    At each prompt, the tracker must either update its prediction or switch targets based on user input. 
    $\checkmark$ and $\times$ denote correct and incorrect responses, respectively.
    }
    \label{fig:protocol}
    \vspace{-4mm}
\end{figure*}

\subsection{Data Acquisition and Scenarios}
To ensure robustness and generalization across real-world applications, we curate a diverse, high-quality video collection spanning six representative scenarios, each characterized by distinct challenges.
Videos are sourced from established public datasets (\eg, LaSOT~\cite{lasot}, VastTrack~\cite{vasttrack}, TNL2K~\cite{tnl2k}, SportsMOT~\cite{sportsmot}, VOT~\cite{vot2020}, LaSOT$_{ext}$~\cite{lasot_ext}, YouTube-VOS~\cite{youtube_vos} and additional online repositories).
All sequences are re-annotated following our interactive annotation protocol, and no existing labels are reused.

The benchmark contains 150 videos with over 140K frames, averaging 947 frames per sequence (see Tab.~\ref{tab:benchmarks_comparison}).
This scale achieves a balanced trade-off between evaluation comprehensiveness and the feasibility of dense, high-quality interactive annotations.
The dataset covers six scenarios:
\begin{itemize}
\item \textbf{Daily Activities:} Everyday personal and household activities from third-person viewpoints with moderate motion, such as walking a dog indoors or in a park.
\item \textbf{Sports Analysis:} Fast motion, frequent occlusions, and visually similar targets requiring dynamic focus shifts, as in soccer, tennis, or basketball scenes.
\item \textbf{UAV Tracking:} Aerial views with large-scale variation, frequent perspective changes, and low-resolution targets, such as vehicles tracked from drones.
\item \textbf{Surveillance:} Crowded public environments with multiple potential targets and challenging lighting, including intersections, parking lots, and subway stations.
\item \textbf{Wildlife Monitoring:} Complex target motion and background camouflage, such as tracking animals (\eg, deer) in natural environments.
\item \textbf{Other Scenarios:} Specialized domains such as industrial monitoring (\eg, robotic arms or machinery) and medical procedures involving surgical tools or human hands.
\end{itemize}

\subsection{Annotation}
High-quality annotation is essential for establishing a reliable benchmark.
InteractTrack employs a rigorous multi-stage annotation pipeline to ensure both accuracy and consistency across all label modalities, including bounding boxes, target descriptions, and scenario labels.

\noindent\textbf{Bounding Box Annotation.}
We provide frame-by-frame, axis-aligned bounding boxes that tightly enclose the visible region of the target.
Each bounding box is independently verified by at least two annotators, and any disagreement is resolved through consensus review by a senior annotator.
For small or distant objects, annotators use a zoom-in interface with sub-pixel alignment and visibility aids to maintain high precision.
When the target is only partially visible, the box encloses the remaining visible area as tightly as possible.
If the target becomes fully occluded or leaves the field of view, the frame is assigned an explicit `absent' label instead of a bounding box.
This provides a clear semantic indication that the object is currently unobservable and maintains compatibility across different evaluation protocols.
The annotation system preserves this label consistently until the target reappears.
Finally, all bounding boxes and absence labels undergo automated validation followed by manual spot checks to ensure annotation integrity.

\noindent\textbf{Interactive Language Annotation.}
Complementing the bounding box annotations, InteractTrack introduces interactive language labels that capture user intent and guide adaptive tracking behavior.
While bounding boxes provide spatial precision, language offers contextual and goal-driven instructions, allowing the tracker to interpret how a user expects it to respond.
Each sequence includes four to five natural-language descriptions inserted at critical moments to emulate real user interactions.
Annotations are produced through a human–GPT–human pipeline: initially drafted by annotators, refined for fluency and naturalness, and finally verified for semantic correctness.
The descriptions encompass multiple interaction types, including initialization, drift correction, focus refinement, and intent switching when the user changes the tracking target.
Even when the target is absent, language cues describe its disappearance or reappearance, enabling the benchmark to jointly encode spatial accuracy and interactive understanding.

\subsection{Dataset Statistics}
InteractTrack comprises 150 videos, totaling over 140K frames and more than 700 natural-language descriptions.
Compared with existing tracking evaluation benchmarks, it is uniquely designed for interactive and multi-scenario evaluation.
The dataset spans diverse scenarios, including sports, daily activities, and surveillance, and exhibits substantial variation in scale, occlusion, and motion.
Each video is annotated with frame-level bounding boxes and interaction-aware language cues, supporting both perceptual and response-level evaluation.

As summarized in Tab.~\ref{tab:benchmarks_comparison}, prior benchmarks such as OTB~\cite{otb}, LaSOT~\cite{lasot}, and GOT-10k~\cite{got10k} focus on large-scale visual diversity but lack interaction-oriented supervision.
In contrast, InteractTrack introduces explicit interactive and scenario labels, enabling systematic assessment of a tracker’s ability to interpret user intent, switch targets, and maintain robustness across multiple contexts.
This design bridges the gap between static tracking benchmarks and the growing need for human-in-the-loop visual understanding.

\begin{figure*}[t!]
    \centering
    \includegraphics[width=0.78\linewidth]{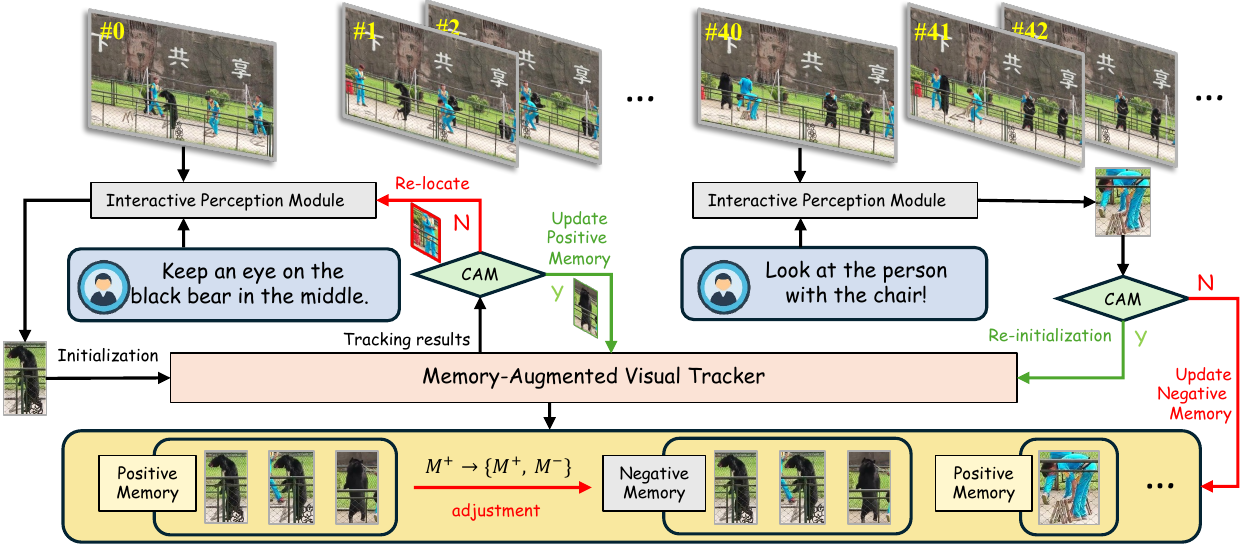}
    \vspace{-2mm}
    \caption{
\textbf{Overview of the proposed Interactive Memory-Augmented Tracking (IMAT) framework.}
The user provides natural-language descriptions of the target to the interactive perception module, which performs reasoning-guided grounding.
The cognitive arbitration module then compares the grounded result with the prediction of the tracker to either correct the trajectory and update the memory banks or confirm the current tracking state.
}
    \label{fig:arch}
    \vspace{-6mm}
\end{figure*}

\subsection{Evaluation Protocol}
Our evaluation protocol is designed to comprehensively assess four complementary aspects of an interactive tracker: Perception, Responsiveness, Tracking, and Interactiveness.
As illustrated in Fig.~\ref{fig:protocol}, each dimension corresponds to a distinct stage of the user–tracker interaction cycle.

Let $B_t$ and $G_t$ denote the predicted and ground-truth bounding boxes at frame $t$, respectively.
For valid frames, the localization accuracy is quantified by the Intersection-over-Union (IoU), defined as $\mathrm{IoU}(B_t, G_t)=\frac{|B_t\cap G_t|}{|B_t\cup G_t|}$.

\noindent\textbf{Perception Capability.}
At frames where a user prompt occurs ($\mathcal{E}$), we assess the tracker’s ability to accurately localize the described target.
Localization is considered correct if the IoU between the predicted and ground-truth bounding boxes exceeds a threshold of $\tau = 0.5$:
\begin{equation}
\label{eq:acc_prec_perc}
\begin{aligned}
\mathrm{Acc}_{\text{perc}}
&= \frac{1}{|\mathcal{E}|}\sum_{t \in \mathcal{E}} \big[\mathrm{IoU}(B_t, G_t) > \tau\big], \\
\mathrm{Prec}_{\text{perc}}
&= \frac{1}{|\mathcal{E}|}\sum_{t \in \mathcal{E}} \mathrm{IoU}(B_t, G_t).
\end{aligned}
\end{equation}
 
\noindent\textbf{Responsiveness.}
When the user issues a prompt to switch targets, the tracker must promptly and accurately track the newly specified object.
A switch is considered correct if the predicted bounding box is closer to the new target $G^{\text{new}}_t$ than to the old one $G^{\text{old}}_t$, with sufficient overlap (IoU $> 0.5$).
Formally, the switch accuracy is defined as:
\begin{equation}
\label{eq:acc_resp}
\mathrm{Acc}_{\text{resp}} =
\frac{1}{|\mathcal{E}_{\text{sw}}|}
\sum_{t \in \mathcal{E}_{\text{sw}}}
\left[\mathrm{IoU}(B_t, G^{\text{new}}_t) >
\mathrm{IoU}(B_t, G^{\text{old}}_t)\right].
\end{equation}

\noindent\textbf{Tracking Capability.}
The intrinsic stability of the tracker is assessed using standard metrics adopted in prior benchmarks such as LaSOT~\cite{lasot} and OTB~\cite{otb}, including Area Under the Curve (AUC) and Precision.

\noindent\textbf{Interactiveness.}
User prompts divide each video into $K$ interaction segments.
For segment $k$ with valid frames $\mathcal{S}_k$, we compute the Interactive Score as follows:
\begin{equation}
\label{eq:interactiveness}
\text{Interactive Score} = 
\frac{1}{K} \sum_{k=1}^{K}
\frac{1}{|\mathcal{S}_k|}
\sum_{t \in \mathcal{S}_k} \mathrm{IoU}(B_t, G_t),
\end{equation}
which measures the overall effectiveness of user–tracker collaboration across all interaction intervals.

Together, these four metrics—Perception, Responsiveness, Tracking, and Interactiveness—provide a holistic evaluation framework for interactive tracking, capturing a system’s ability to perceive, respond, and adapt in user-driven scenarios.

\section{Interactive Memory-Augmented Tracking}
In this section, we introduce Interactive Memory-Augmented Tracking (IMAT), a strong baseline for long-term interactive tracking that integrates the spatiotemporal consistency of visual tracking with the semantic reasoning capabilities of multimodal large language models (MLLMs).
IMAT enables the tracker to learn from user feedback through dynamic memory reasoning, progressively improving robustness, responsiveness, and semantic alignment.
As illustrated in Fig.~\ref{fig:arch}, IMAT comprises three key components—the Interactive Perception Module (IPM), the Memory-Augmented Visual Tracker (MAVT), and the Cognitive Arbitration Module (CAM)—which operate collaboratively to achieve adaptive and semantically user-guided tracking.

\subsection{Interactive Perception Module}
The Interactive Perception Module (IPM) serves as the human-in-the-loop interface, providing semantic guidance to the tracker.
At any frame $t$, the user may issue a natural-language query $P_t$ (\eg, `keep an eye on the black bear in the middle').
The IPM processes both the current frame $I_t$ and the query $P_t$ to perform vision–language grounding, producing a semantically aligned bounding box $B_{\text{ipm}}(t)$.

In our implementation, the IPM is instantiated with an MLLM-based perception model, Rex-Omni~\cite{rex},  which aligns visual features with user-provided descriptions.
Operating on the current frame and user prompt, the IPM outputs a grounded box that can either re-initialize or validate the tracker through the Cognitive Arbitration Module.

\subsection{Memory-Augmented Visual Tracker}
The Memory-Augmented Visual Tracker (MAVT) builds upon SAM2~\cite{sam2}, which maintains spatiotemporal consistency across frames.
We extend SAM2 with two external memory banks—a positive memory $\mathcal{M}^+$ and a negative memory $\mathcal{M}^-$—that enable adaptive appearance learning and distractor suppression.
At each frame $t$, the tracker predicts a bounding box conditioned on both memory banks:
\begin{equation}
\label{eq:tracker}
B_{\text{track}}(t) = \mathrm{Tracker}\big(I_t\,;\,\mathcal{M}^+,\,\mathcal{M}^-\big).
\end{equation}
The positive memory $\mathcal{M}^+$ stores embeddings of verified target cues, enabling the tracker to adapt to valid variations in pose, illumination, and scale over time.
Conversely, the negative memory $\mathcal{M}^-$ retains embeddings of distractors, failed predictions, or switched targets, helping the tracker suppress responses in previously confused or ambiguous regions.
Both memories are dynamically updated under novelty and diversity constraints to maintain a compact yet expressive representation.
This adaptive update policy, implemented on top of SAM2, ensures continuous adaptation while preventing memory redundancy.

\begin{table*}[t]
\centering
\caption{\textbf{Combined evaluation on the InteractTrack test set under the same interactive protocol.} Methods are grouped by paradigm (VLT/VOS/VOT/RVOS); best results in each column are in \textbf{bold}.}
\vspace{-2mm}
\label{tab:combined_tracking_results}
\renewcommand{\arraystretch}{0.8}
\resizebox{\linewidth}{!}{%
\begin{NiceTabular}{llrrrrrrr}
\CodeBefore
\rowcolors{3}{gray!10}{white}
\columncolor{white}{1}
\Body
\toprule
\Block{2-1}{\textbf{Type}} & \Block{2-1}{\textbf{Algorithm}} & \Block{2-1}{\textbf{Interactiveness}} & \Block{2-1}{\textbf{Responsiveness}} & \Block{1-2}{\textbf{Perception Capability}} & & \Block{1-3}{\textbf{Tracking Capability}} \\
\cmidrule(lr){5-6} \cmidrule(lr){7-9}
& & & & Accuracy  & Precision & AUC & Precision & Norm Precision \\
\midrule
& \textbf{Ours} & \textbf{45.25} & \textbf{41.20} & \textbf{52.78} & \textbf{49.63} & \textbf{45.86} & \textbf{49.63} & \textbf{60.90} \\
\midrule
\Block{3-1}{VLT}
& JointNLT~\cite{zhou2023joint} & 30.66 & 36.67 & 44.33 & 43.08 & 19.81 & 16.16 & 30.44 \\
& DUTrack~\cite{li2025dynamic}  & 39.52 & 37.83 & 48.82 & 47.09 & 41.70 & 44.85 & 56.55 \\
& SUTrack~\cite{sutrack}  & 40.90 & 38.04 & 49.25 & 48.38 & 44.25 & 47.23 & 58.26 \\
\midrule
\Block{5-1}{VOS}
& SAM2~\cite{sam2}     & 41.43 & 35.93 & 47.43 & 45.58 & 42.22 & 45.12 & 56.42 \\
& HIM2SAM~\cite{him2sam}  & 42.72 & 37.83 & 49.68 & 46.10 & 41.39 & 46.10 & 57.16 \\
& SAMURAI~\cite{samurai}  & 43.69 & 37.20 & 49.36 & 46.44 & 41.53 & 45.57 & 56.59 \\
& VL-SAM2~\cite{vlsam2}  & 44.43 & 37.72 & 48.82 & 46.52 & 41.88 & 45.73 & 56.84 \\
& DAM4SAM~\cite{dia4sam}  & 43.19 & 37.62 & 49.89 & 46.58 & 43.79 & 48.74 & 59.72 \\
\midrule
\Block{15-1}{VOT}
& ToMP~\cite{tomp}     & 35.92 & 37.93 & 46.79 & 44.95 & 36.14 & 37.46 & 50.63 \\
& SeqTrack~\cite{seqtrack} & 37.61 & 37.51 & 46.15 & 45.45 & 40.89 & 44.43 & 55.64 \\
& TaMOs~\cite{tamos}    & 36.56 & 37.72 & 47.86 & 46.33 & 36.16 & 38.80 & 50.01 \\
& SimTrack~\cite{simtrack} & 36.88 & 37.62 & 47.00 & 45.91 & 37.71 & 38.33 & 51.22 \\
& CiteTrack~\cite{citetracker} & 37.37 & 38.36 & 46.57 & 45.25 & 37.74 & 39.57 & 51.46 \\
& STARK~\cite{stark}    & 36.54 & 38.57 & 49.36 & 46.58 & 35.37 & 36.31 & 49.28 \\
& MixFormer~\cite{mixformer} & 37.13 & 37.62 & 47.75 & 46.37 & 40.39 & 42.49 & 54.65 \\
& OSTrack~\cite{ostrack}  & 37.65 & 38.57 & 48.07 & 46.14 & 38.46 & 40.62 & 53.05 \\
& DropTrack~\cite{droptrack} & 38.04 & 38.15 & 46.90 & 45.61 & 39.02 & 41.02 & 53.38 \\
& ARTrack~\cite{artrack}  & 38.73 & 38.36 & 46.57 & 45.82 & 41.11 & 43.75 & 55.71 \\
& GRM~\cite{tracker_grm}     & 38.31 & 38.46 & 48.82 & 47.12 & 39.39 & 40.99 & 53.42 \\
& ROMTrack~\cite{ROMTrack} & 39.05 & 38.15 & 48.39 & 47.12 & 41.29 & 43.93 & 56.08 \\
& ODTrack~\cite{odtrack} & 39.34 & 38.46 & 49.14 & 47.71 & 43.30 & 46.69 & 58.87 \\
& MCITrack~\cite{mcitrack} & 40.38 & 37.93 & 47.97 & 47.48 & 44.98 & 47.92 & 59.61 \\
\midrule
\Block{2-1}{RVOS}
& VideoLiSA~\cite{videolisa} & 39.82 & 32.67 & 37.90 & 39.99 & 22.77 & 16.73 & 29.30 \\
& Sa2VA~\cite{sa2va}     & 44.81 & 38.99 & 45.50 & 46.05 & 24.14 & 21.10 & 33.39 \\
\bottomrule
\end{NiceTabular}
}
\vspace{-6mm}
\end{table*}

\subsection{Cognitive Arbitration Module}
The Cognitive Arbitration Module (CAM) governs the interaction between the IPM and MAVT, acting as a high-level decision controller that determines whether to maintain or correct the tracker state.
It is activated at interactive frames whenever a user prompt triggers the IPM or the tracker detects motion inconsistency.
At an interactive frame $t$, the CAM compares the tracker’s prediction $B_{\text{track}}(t)$ with the grounded box from the IPM $B_{\text{ipm}}(t)$ using the Intersection-over-Union (IoU):
\vspace{-2mm}
\begin{equation}
\label{eq:iou}
\mathrm{IoU} =
\frac{\mathrm{area}(B_{\text{track}}(t) \cap B_{\text{ipm}}(t))}
{\mathrm{area}(B_{\text{track}}(t) \cup B_{\text{ipm}}(t))}.
\end{equation}
A potential drift is detected when the IoU drops below a threshold $\tau_{\text{iou}}$ or when the displacement between consecutive box centers exceeds $\delta_c$.
In such cases, the CAM invokes the IPM to re-verify whether the current prediction corresponds to the intended target.
If the grounding result indicates a mismatch, the tracker is re-initialized using $B_{\text{ipm}}(t)$, and the memory banks are updated accordingly: the failed embedding is added to the negative memory $\mathcal{M}^-$, while the corrected one is added to the positive memory $\mathcal{M}^+$.
Otherwise, when the two outputs are consistent, the tracker continues propagation, and $\mathcal{M}^+$ is reinforced with the current feature embedding.
The final prediction at frame $t$ is then defined as:
\vspace{-2mm}
\begin{equation}
\label{eq:final_box}
B_{\text{final}}(t) =
\begin{cases}
B_{\text{ipm}}(t), & \text{if drift/mismatch detected},\\
B_{\text{track}}(t), & \text{otherwise.}
\end{cases}
\end{equation}
In practice, we set $\tau_{\text{iou}}^{\text{init}} = 0.3$ during initialization to avoid overlap and $\tau_{\text{iou}}^{\text{reinit}} = 0.6$ for runtime arbitration.
This selective arbitration mechanism integrates spatial, semantic, and motion cues only when necessary, ensuring efficient, robust, and stable interactive tracking.

\subsection{Overall Interaction Process}
As illustrated in Fig.~\ref{fig:arch}, the IMAT pipeline begins with user initialization, followed by continuous tracking through the MAVT.
During tracking, users can issue interaction prompts at arbitrary frames to provide updated natural-language descriptions.
Each interaction is processed by the IPM for semantic grounding, while the CAM governs when and how user feedback influences the tracker.
When the CAM detects divergence, the MAVT is re-initialized using the grounded result; otherwise, its state and memory banks are reinforced.
This bidirectional update mechanism—comprising positive reinforcement for consistent frames and negative learning for failures—enables IMAT to continuously learn from user interaction, resulting in robust, adaptive, and interpretable long-term tracking.


\section{Benchmarking}
\subsection{Evaluated Trackers}
We evaluate four tracker groups on the InteractTrack benchmark: VLT, VOS, VOT, and RVOS.
This selection encompasses a broad spectrum of visual and multimodal architectures, ranging from traditional single-object trackers to language-grounded models.
VLT trackers (JointNLT, SUTrack, DUTrack) initialize from template descriptions and update the reference whenever a prompt arrives. 
VOS (SAM2, VL-SAM2, SAMURAI, DAM4SAM, HIM2SAM) start from the first-frame box; VL-SAM2 uses an external grounding module, and all mask outputs are converted to boxes for scoring.
VOT (MCITrack, ODTrack, ROMTrack, MixViT, GRM, ARTrack, DropTrack, OSTrack, MixFormer, STARK, CiteTrack, SimTrack, TaMOS, SeqTrack, TOMP) receive only the first-frame box and no text; on each new prompt, we re-initialize them with the corresponding ground-truth box to isolate tracking ability.
RVOS (Sa2VA, VideoLISA) are initialized and updated from text, with masks being converted to tight boxes accordingly. 
All methods are evaluated under the same interactive schedule, with identical prompt timing and initialization policy, ensuring a unified comparison across model categories.

\begin{figure*}[t]
\centering
\begin{minipage}{0.275\linewidth} 
    \centering
    \includegraphics[width=\linewidth]{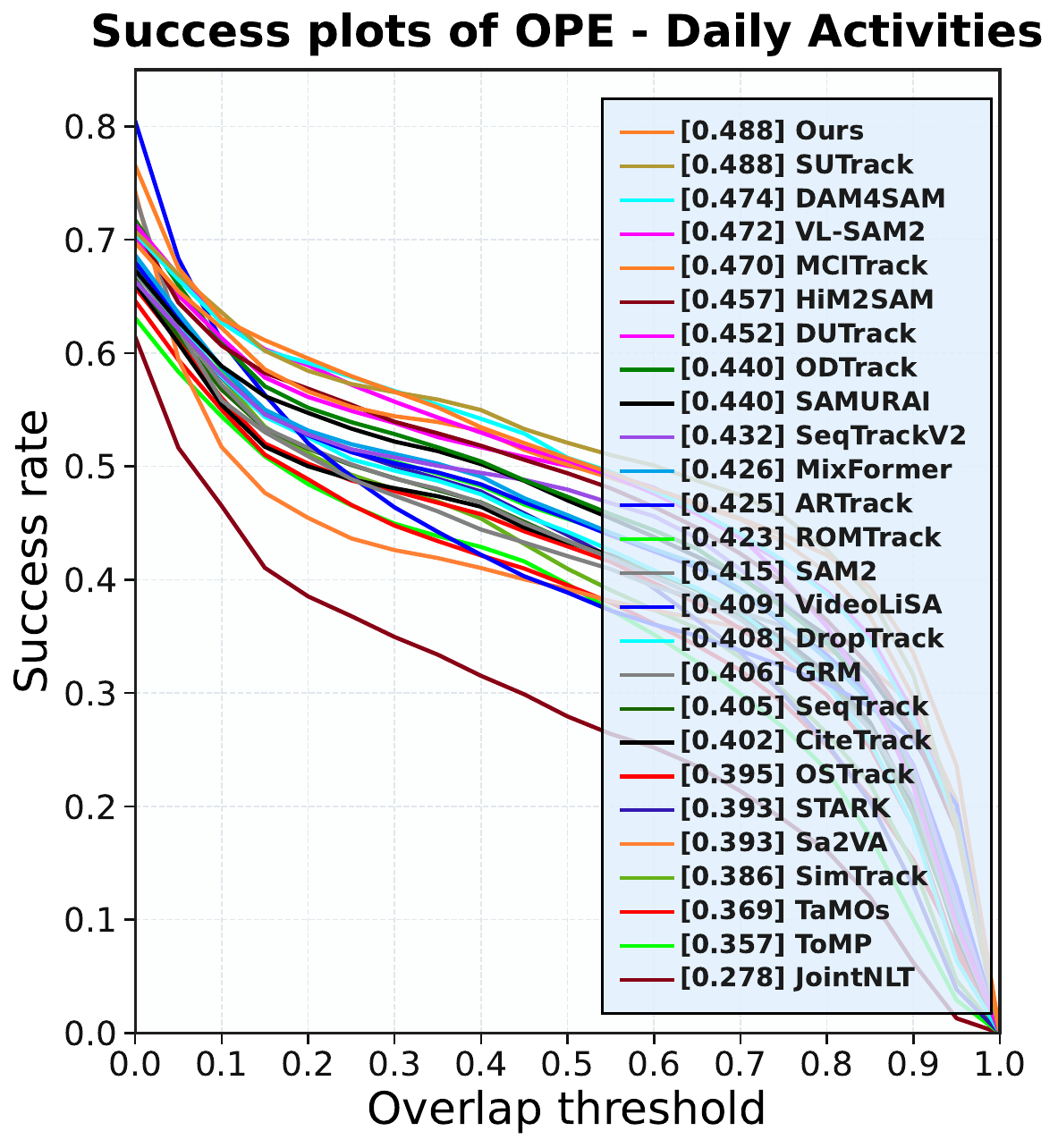}
\end{minipage}
\hspace{0.04\linewidth}
\begin{minipage}{0.28\linewidth}
    \centering
    \includegraphics[width=\linewidth]{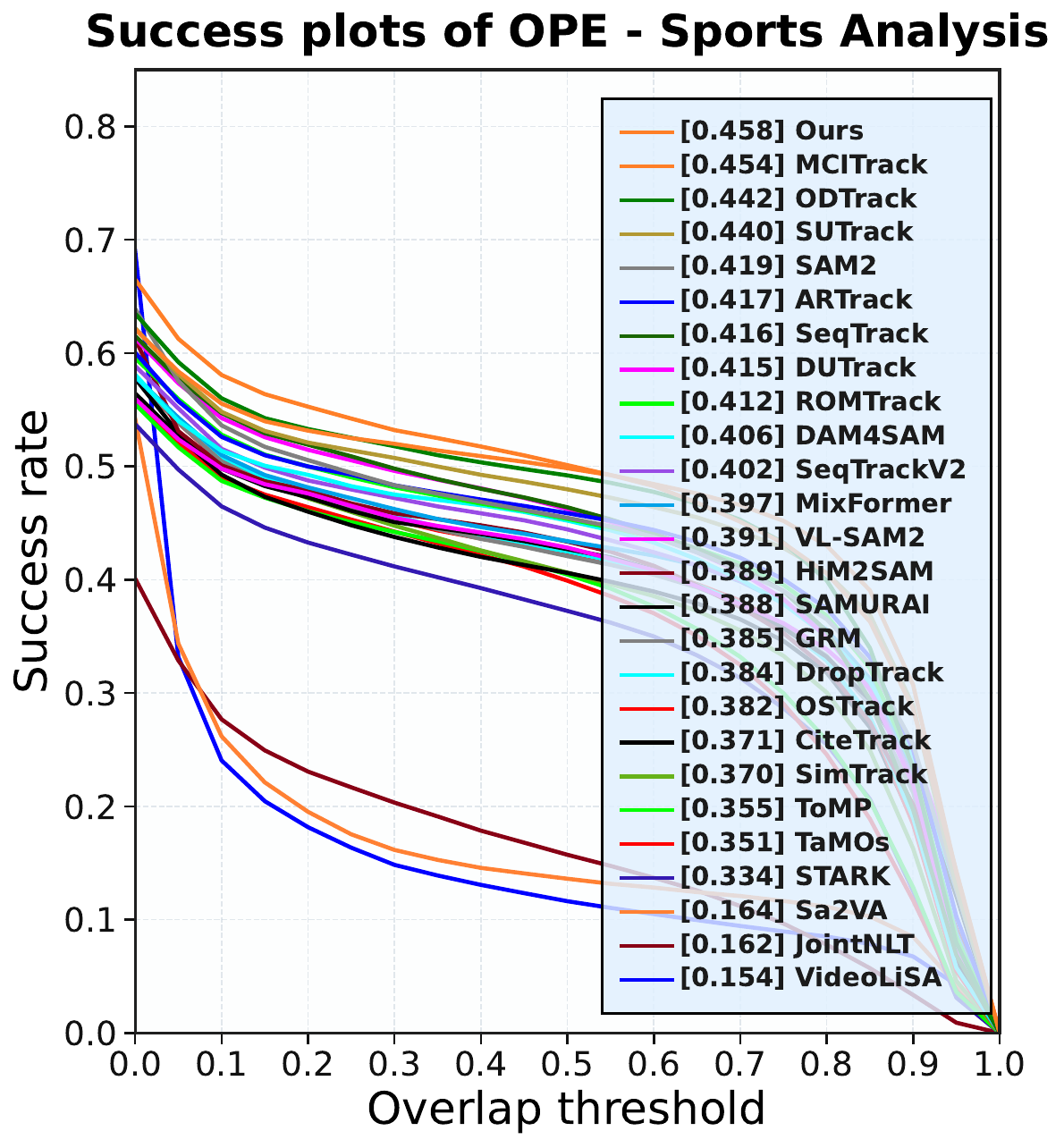}
\end{minipage}%
\hspace{0.04\linewidth}
\begin{minipage}{0.27\linewidth}
    \centering
    \includegraphics[width=\linewidth]{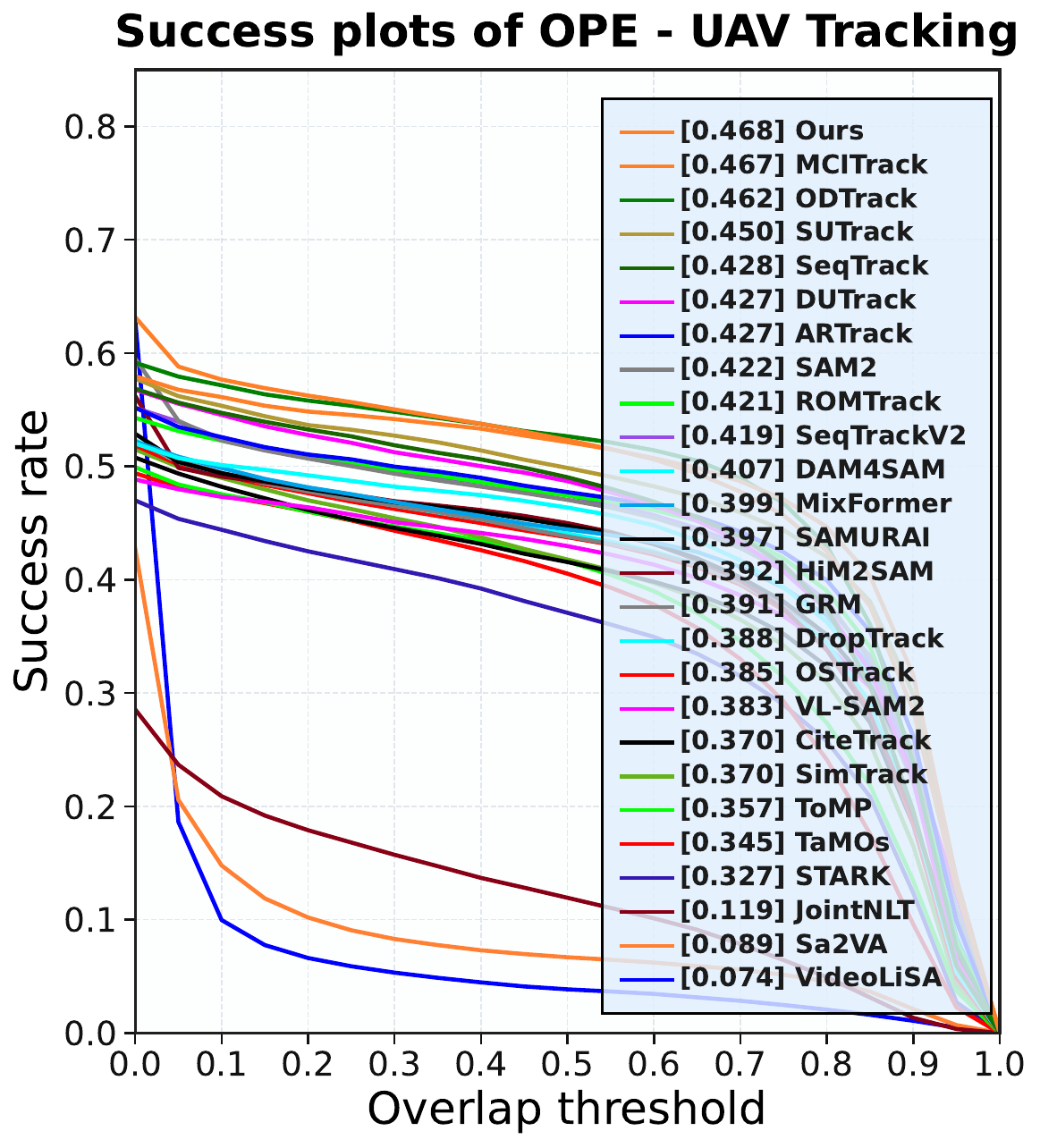}
\end{minipage}

\vspace{-1ex} 

\begin{minipage}{0.28\linewidth}
    \centering
    \includegraphics[width=\linewidth]{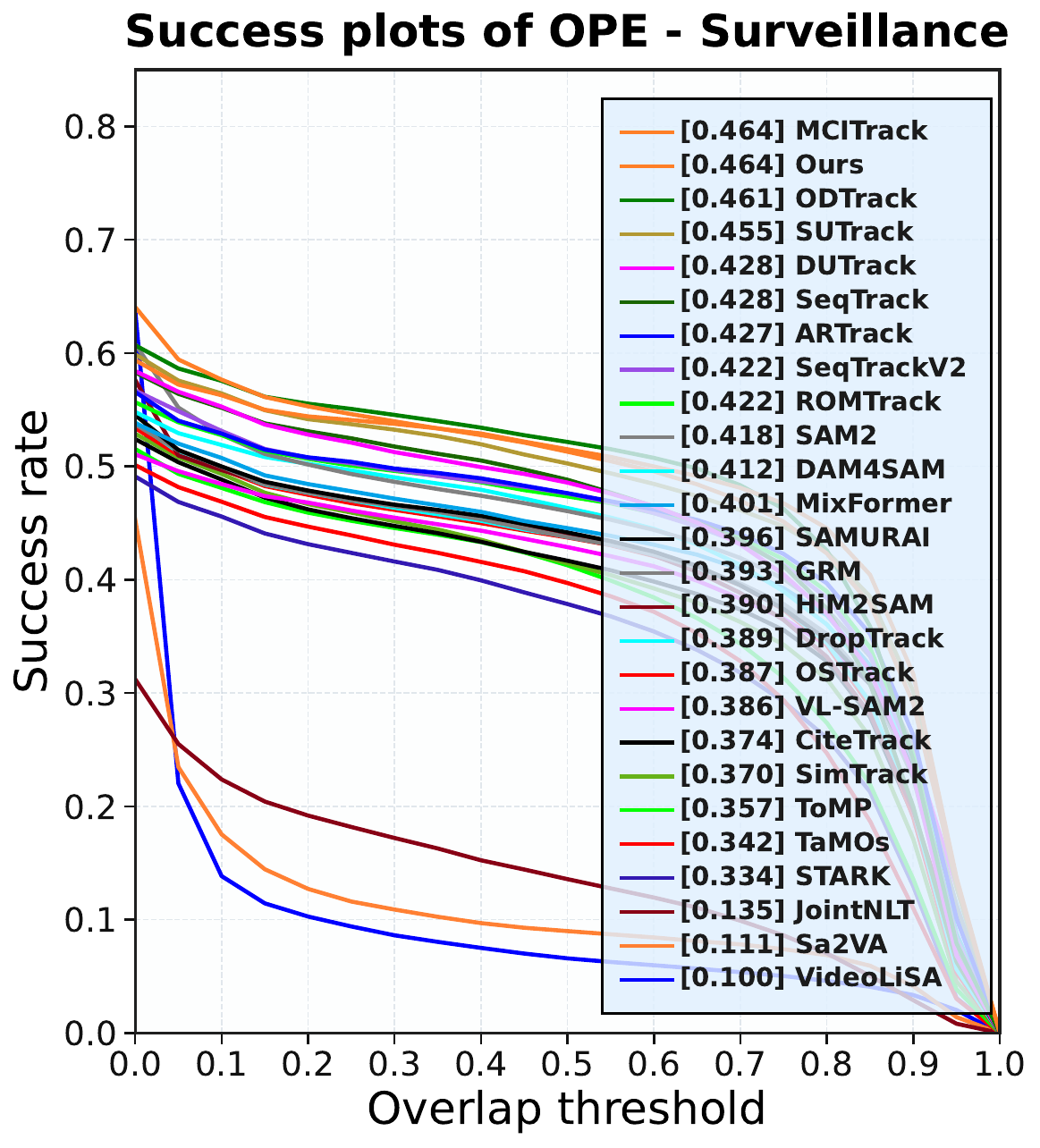}
\end{minipage}%
\hspace{0.04\linewidth}%
\begin{minipage}{0.28\linewidth}
    \centering
    \includegraphics[width=\linewidth]{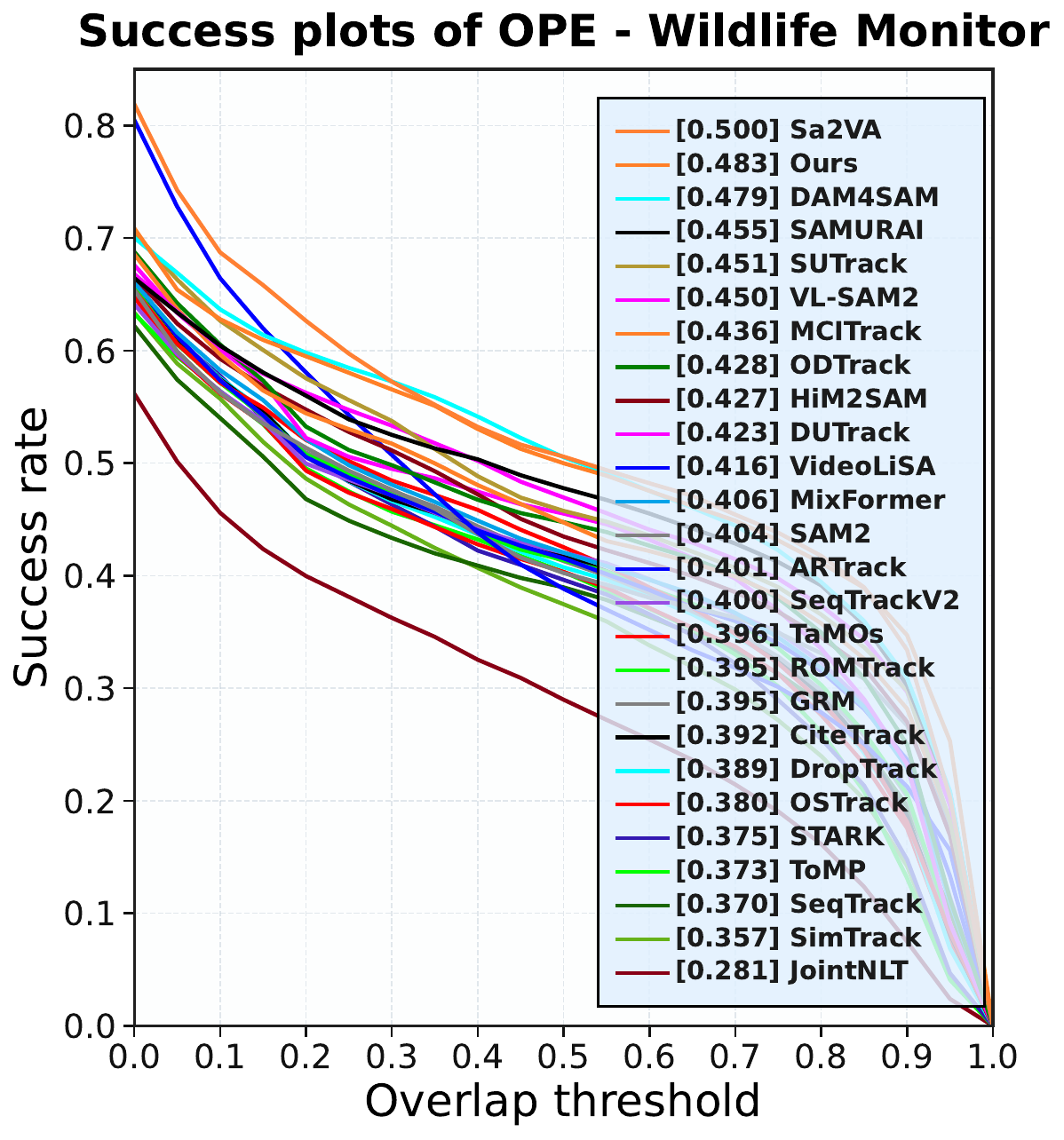}
\end{minipage}%
\hspace{0.04\linewidth}%
\begin{minipage}{0.28\linewidth}
    \centering
    \includegraphics[width=\linewidth]{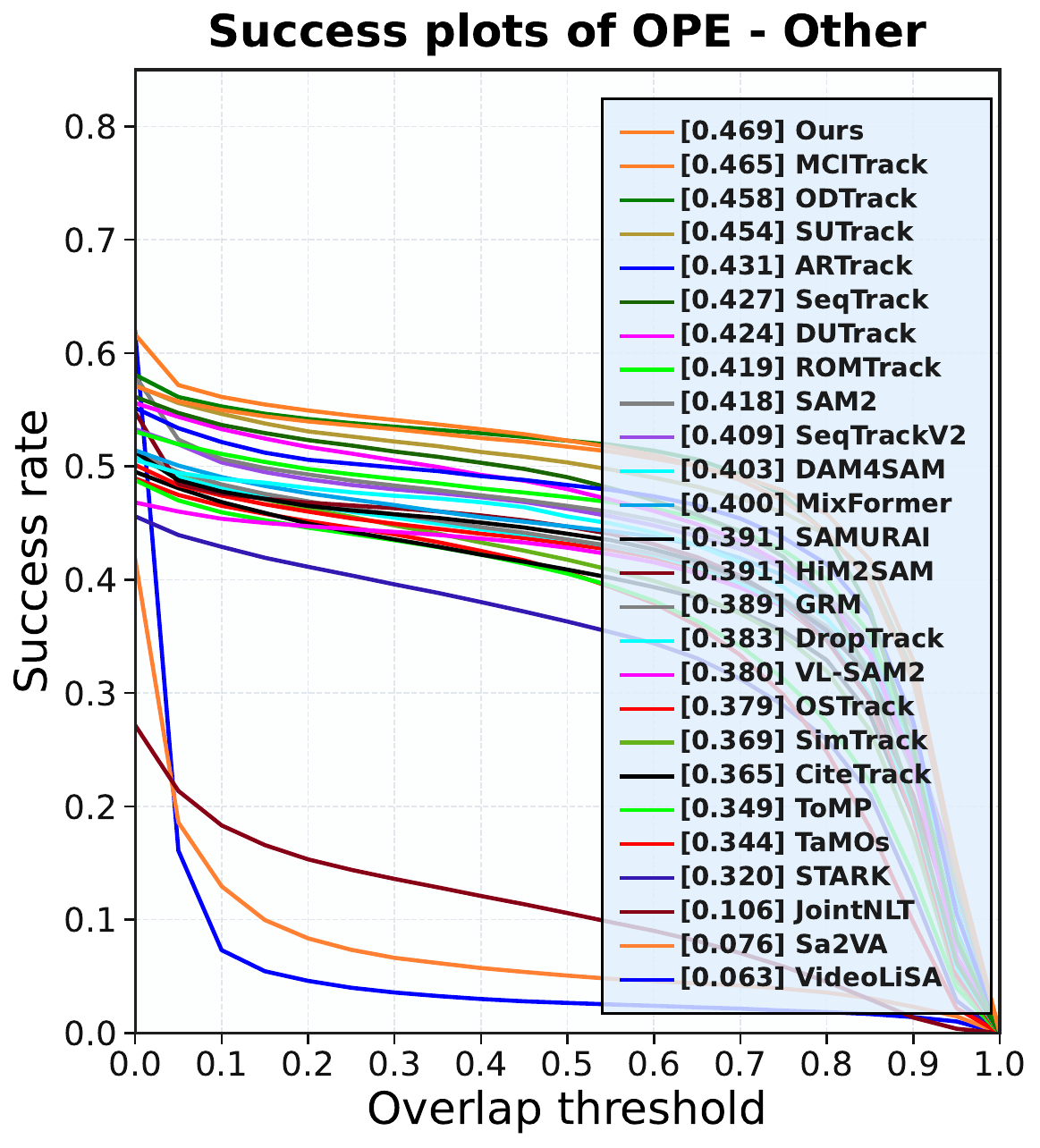}
\end{minipage}

\vspace{-4mm}
\caption{\textbf{Evaluation results on the six scenarios using the success rate in the InteractTrack benchmark.} These figures show that the proposed baseline performs strongly across all six scenarios.}
\label{fig:scenarios}
\vspace{-4mm} 
\end{figure*}
\subsection{Evaluation Results}
Tab.~\ref{tab:combined_tracking_results} presents a comprehensive comparison of all evaluated trackers across four key dimensions: Interactiveness, Responsiveness, Perception, and Tracking Capability.
Our proposed method achieves the best performance, demonstrating accurate target localization, fast adaptation to user prompts, and stable long-term tracking.
Specifically, it attains the highest Interactiveness Score (45.25\%) and Responsiveness (41.20\%), indicating a stronger understanding of natural-language instructions and more effective adaptation to interactive cues.
Among the baselines, VLT models such as SUTrack and DUTrack achieve relatively high responsiveness and perception accuracy due to their joint vision–language representations.
However, their localization tends to degrade over longer sequences, revealing limited temporal robustness.
VOS-based methods (\eg, SAMURAI, VL-SAM2) benefit from segmentation priors and deliver strong short-term accuracy, yet their mask-based tracking is easily disrupted by occlusion and rapid motion.
VOT trackers (\eg, MixViT, OSTrack, STARK) maintain reliable precision under static or well-constrained conditions but cannot interpret or react to textual guidance, resulting in weak performance in interactive scenarios.
Finally, RVOS approaches can respond to textual cues but exhibit poor temporal stability and reduced robustness during continuous user interaction.
Overall, these findings show that while existing trackers perform competitively in conventional autonomous settings, they struggle to generalize to dynamic, user-driven tasks.
By jointly integrating perception, tracking, and interaction, our method achieves consistent improvements across all key evaluation dimensions.
These results underscore the importance of InteractTrack as a benchmark for advancing adaptive, human-in-the-loop visual tracking systems.

\subsection{Scenario-based Performance}
Fig.~\ref{fig:scenarios} presents scenario-level success plots under the One-Pass Evaluation (OPE) protocol across six representative scenarios: daily activities, sports analysis, UAV tracking, surveillance, wildlife monitoring, and other scenarios.
Our proposed method achieves the overall best performance across most environments, demonstrating strong generalization to diverse motion patterns and scenario dynamics.
In daily activities and sports analysis, where complex human–object interactions and frequent target switches occur, IMAT maintains stable localization and outperforms both vision–language and single-object baselines by a clear margin.
In UAV tracking and surveillance scenes, our model demonstrates robustness to scale variation, camera motion, and long-term viewpoint shifts, surpassing segmentation-based trackers that often drift under large appearance changes.
In wildlife monitoring, where targets are typically small, camouflaged, or partially occluded, performance remains competitive, indicating strong adaptability to low-contrast and irregular motion conditions.
Overall, the consistent superiority of our tracker across all scenario types verifies its capability to perceive, follow, and adapt in dynamic real-world scenarios, validating the effectiveness of the proposed interaction-aware architecture and its generalization across heterogeneous domains.

\begin{table}[tp]
\centering
\caption{\textbf{Ablation study on the InteractTrack benchmark.}}
\vspace{-3mm}
\label{tab:ablation}
\resizebox{\linewidth}{!}{
\begin{tabular}{lcc}
\toprule 
\textbf{Model Variant} & \textbf{Responsiveness (\%)} & \textbf{Interactiveness (\%)} \\
\midrule
w/o IPM (no language) & 35.93 & 41.43 \\
w/o Memory & 37.72 & 41.52  \\
w/o CAM & 38.03 & 42.09 \\
w/ Naive IoU Arbitration & 39.55 & 43.02 \\
\textbf{IMAT (ours)} & \textbf{41.20} & \textbf{45.25} \\
\bottomrule
\end{tabular}
}
\vspace{-6mm}
\end{table}

\subsection{Ablation Studies}
We evaluate the contribution of each IMAT component on the InteractTrack benchmark.
As summarized in Tab.~\ref{tab:ablation}, removing any key module results in a consistent degradation in both Responsiveness and Interactiveness.
Without the Interactive Perception Module (IPM), the tracker loses the ability to interpret user prompts and thus fails to realign after target switches.
Eliminating the memory mechanism prevents adaptive learning of appearance dynamics and distractor suppression, leading to unstable long-term behavior.
Removing the Cognitive Arbitration Module (CAM) causes frequent drift or unnecessary re-initializations; while a naive IoU-based arbitration provides partial recovery, it introduces jitter and redundant corrections.
In contrast, the proposed IMAT—combining semantic grounding, adaptive memory, and reasoning-based arbitration—achieves the best balance between responsiveness and interaction stability.
These results confirm that all three components play complementary roles in enabling robust, user-guided, and adaptive tracking under dynamic conditions.

\section{Conclusion}
We introduce Interactive Visual Tracking, a new paradigm that reflects real-world scenarios where user intent evolves over time.
By incorporating natural-language guidance, a tracker can recover from drift, switch targets, and dynamically adjust its focus during online operation.
To enable systematic study of this problem, we construct InteractTrack, the first large-scale benchmark featuring dense visual and linguistic annotations, and propose Interactive Memory-Augmented Tracking (IMAT), a strong baseline that unifies visual continuity, semantic grounding, and adaptive memory reasoning.

\section{Acknowledgments}
\label{sec:Acknowledgments}
This work was supported in part by the National Natural Science Foundation of China (Grant No.62476148) and the Guangdong Basic and Applied Basic Research Foundation (Grant No.2024A1515011292).

{
    \small
    \bibliographystyle{ieeenat_fullname}
    \bibliography{main}
}

\clearpage
\clearpage
\setcounter{page}{1}
\maketitlesupplementary
\appendix

We provide additional details for the InteractTrack benchmark, including a thorough explanation of its motivation and construction protocol, the interaction dynamics that characterize the task, and scene-wise visualizations covering all six scenarios. Moreover, we present extended experimental results—encompassing precision analysis, challenging-case evaluations, and failure-case studies—that offer deeper insight into tracker behavior under complex and dynamic interaction conditions.

\section{Additional Dataset Details}

\subsection{Motivation and Protocols}

\noindent\textbf{Motivation.}
Existing tracking benchmarks such as LaSOT~\cite{lasot}, GOT-10k~\cite{got10k}, and TrackingNet~\cite{trackingnet} assume a fully autonomous paradigm: the tracker is initialized once and must follow the target throughout the video without further human feedback. While this setting has enabled significant progress, it does not reflect real-world usage, where users frequently correct drift, refine the focus of tracking, or switch targets altogether.

In addition, existing benchmarks impose a fixed target throughout the entire sequence and provide no mechanism to model or evaluate such interactive behaviors. Vision–language tracking datasets (e.g., TNL2K~\cite{tnl2k}) allow natural-language initialization but remain strictly one-shot, offering no support for evolving or time-varying user intent. Multimodal datasets such as MGIT~\cite{mgit} introduce long videos and rich semantic annotations, yet their language descriptions serve as static global context rather than dynamic, incremental instructions.

Consequently, neither autonomous visual tracking nor language-initialized tracking captures the essential dimension of interaction. To bridge this gap, our benchmark is designed to explicitly model the missing interactive component—where users issue instructions over time, and trackers are expected to perceive, interpret, and adapt to these changing directives.

\noindent\textbf{Protocol.}
InteractTrack is designed to simulate real-world scenarios in which human input plays an essential role in the tracking process. We intentionally select sequences that naturally create opportunities for user intervention, including severe occlusions, abrupt appearance changes, ambiguous object interactions, and long-term tracking segments that require re-detection via language cues. The videos span a diverse set of domains—sports, surveillance, UAV footage, daily activities, and wildlife—capturing a broad range of motion patterns and scene complexities.

Each video is annotated with dense, frame-level bounding boxes and timestamped language instructions marking explicit interaction points (e.g., initialization, drift correction, and target switching). To rigorously evaluate a tracker's ability to switch targets while maintaining situational awareness, we annotate both the new target and the previous target at every switch. This dual annotation provides clear ground truth for assessing whether the tracker transitions to the correct entity while appropriately rejecting distractors.

\subsection{Interaction Dynamics}

InteractTrack explicitly models the dynamic interaction process between the user and the tracker. Unlike traditional tracking tasks that assume a fixed target throughout the entire video, our dataset introduces multiple interaction points where human input is required. These interactions are driven by the evolving visual context and include:

\begin{itemize}
\item Changes in target appearance (e.g., a player switching roles or taking possession of the ball),
\item Drift or loss due to occlusion (e.g., a player temporarily disappearing behind others),
\item Role transitions (e.g., when a player passes the ball to a teammate),
\item Shifts in user focus (e.g., switching from tracking the player to tracking the ball).
\end{itemize}

Each interaction point is accompanied by precise ground-truth annotations, with special care given to target-switching events. At every switch, we annotate both the new target and the previous target, enabling rigorous evaluation of whether the tracker transitions correctly while avoiding distractors. Between interaction points, the tracker is expected to operate autonomously, maintaining stable tracking until the next user instruction.

This protocol reflects realistic human-in-the-loop tracking scenarios and allows us to evaluate how well a tracker adapts to corrections, target switches, and evolving user guidance across multiple phases of interaction.

\subsection{Scene-wise Sequence Visualizations}

For each of the six scenarios in our dataset, we include three representative sequences (18 in total) in the supplementary material. Each scenario provides one sequence with detailed textual interaction, explaining the annotation context and highlighting the key qualitative aspects of the scene.

The remaining sequences are provided as visual-only examples, showcasing appearance diversity, intra-scenario variation, and overall scene complexity. Together, these visualizations offer a comprehensive overview of the environments and tracking challenges encompassed by the dataset.

\begin{figure*}[t]
    \centering
    \includegraphics[width=1.0\linewidth]{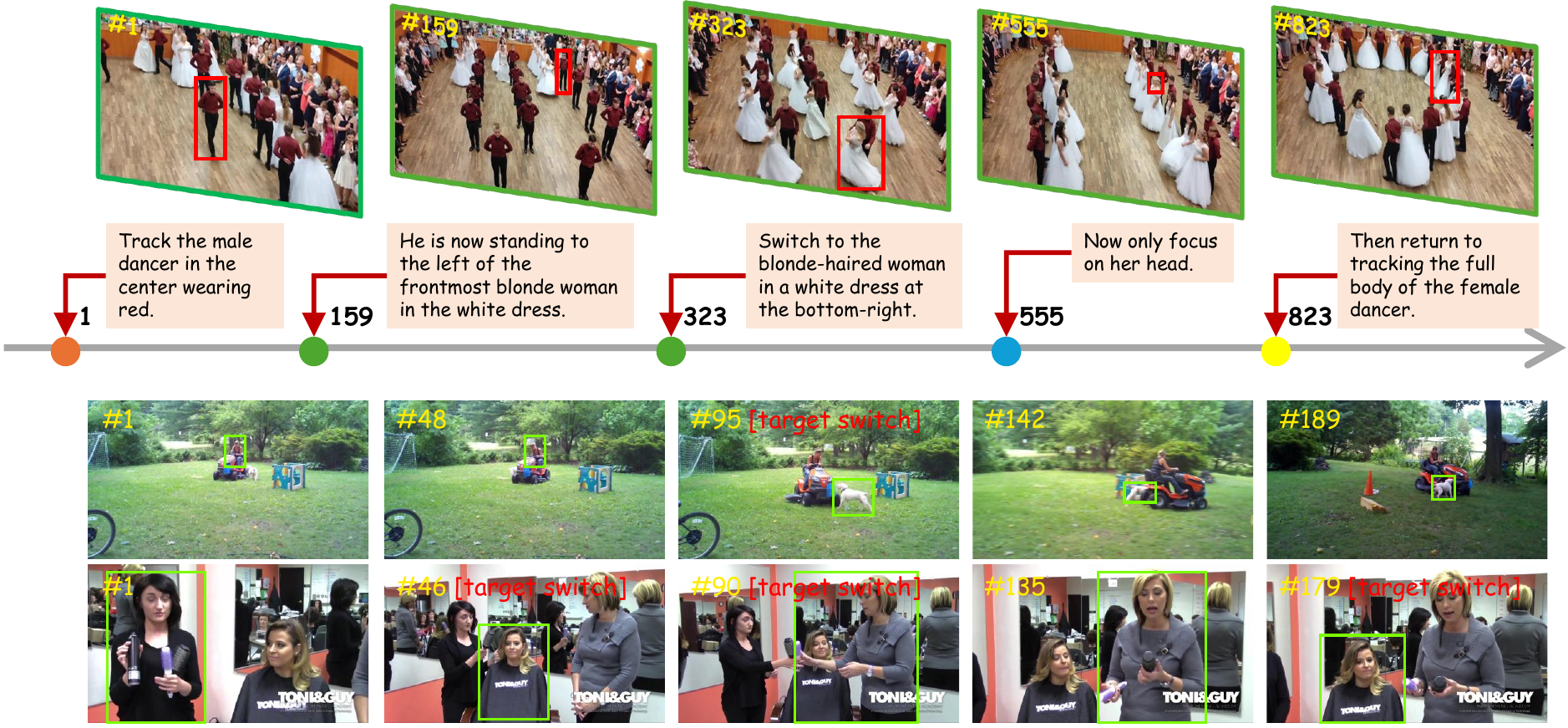}
\caption{
Representative sequences from the \textbf{daily activities} scenario.
This scenario includes everyday indoor and outdoor activities captured from a third-person viewpoint. The sequences depict interactions such as playing with pets, handling household objects, and casual human motion under moderate viewpoint shifts. The highlighted sequence illustrates target-switch interaction cues.
}
    \label{fig:daily_activities}
\end{figure*}

\begin{figure*}[ht]
    \centering
    \includegraphics[width=1.0\linewidth]{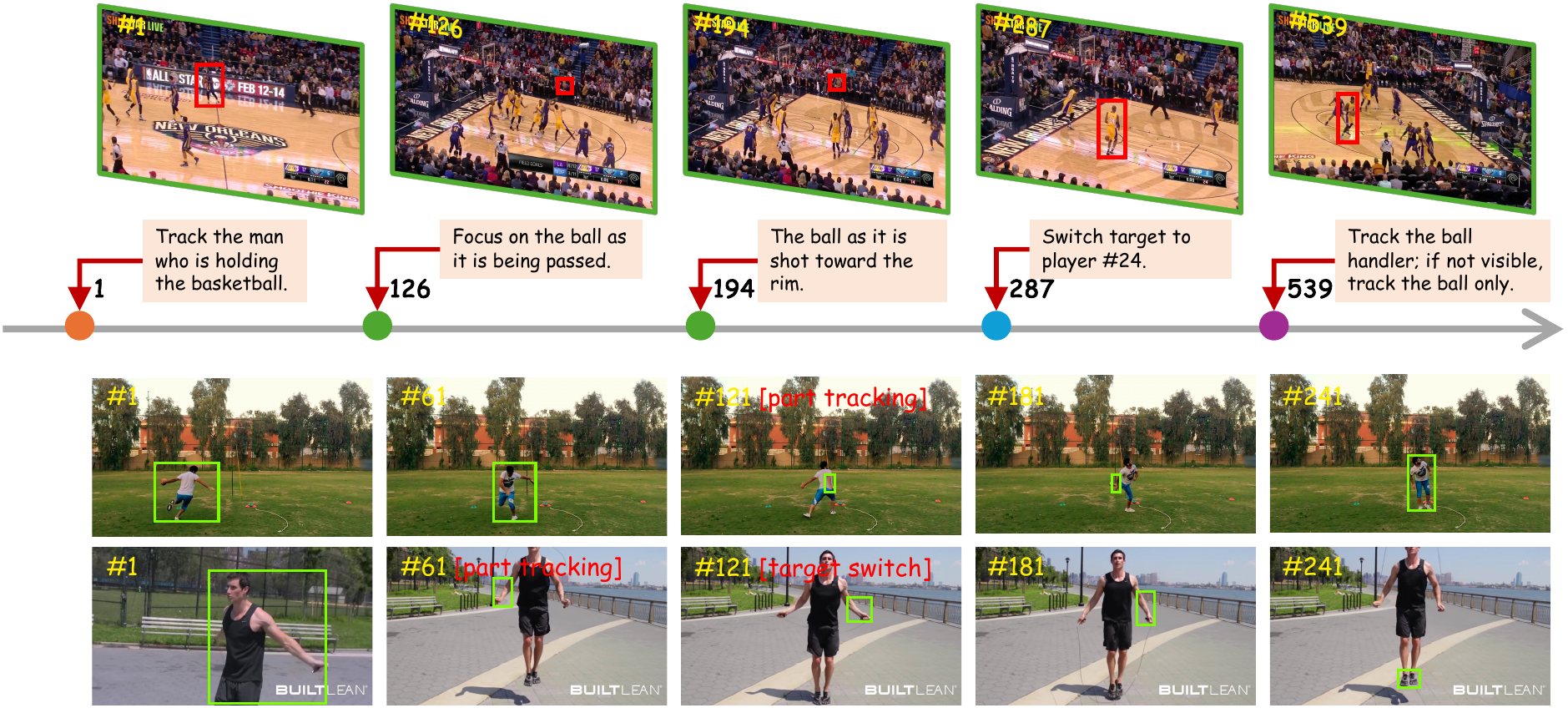}
\caption{
Representative sequences from the \textbf{sports analysis} scenario.
The sequence illustrates multiple interaction points in a basketball game, including target initialization, ball-focused attention during passes, shot-following, and explicit target switching among players. At each key timestamp, the user provides updated instructions that reflect the evolving visual context—shifting attention from the ball handler to the ball itself, following the shot trajectory, and then redirecting focus to a specific player. The highlighted sequence illustrates target-switch and part-tracking interaction cues as user instructions evolve across multiple stages.
}
    \label{fig:sports_analysis}
\end{figure*}

\begin{figure*}[ht]
    \centering
    \includegraphics[width=1.0\linewidth]{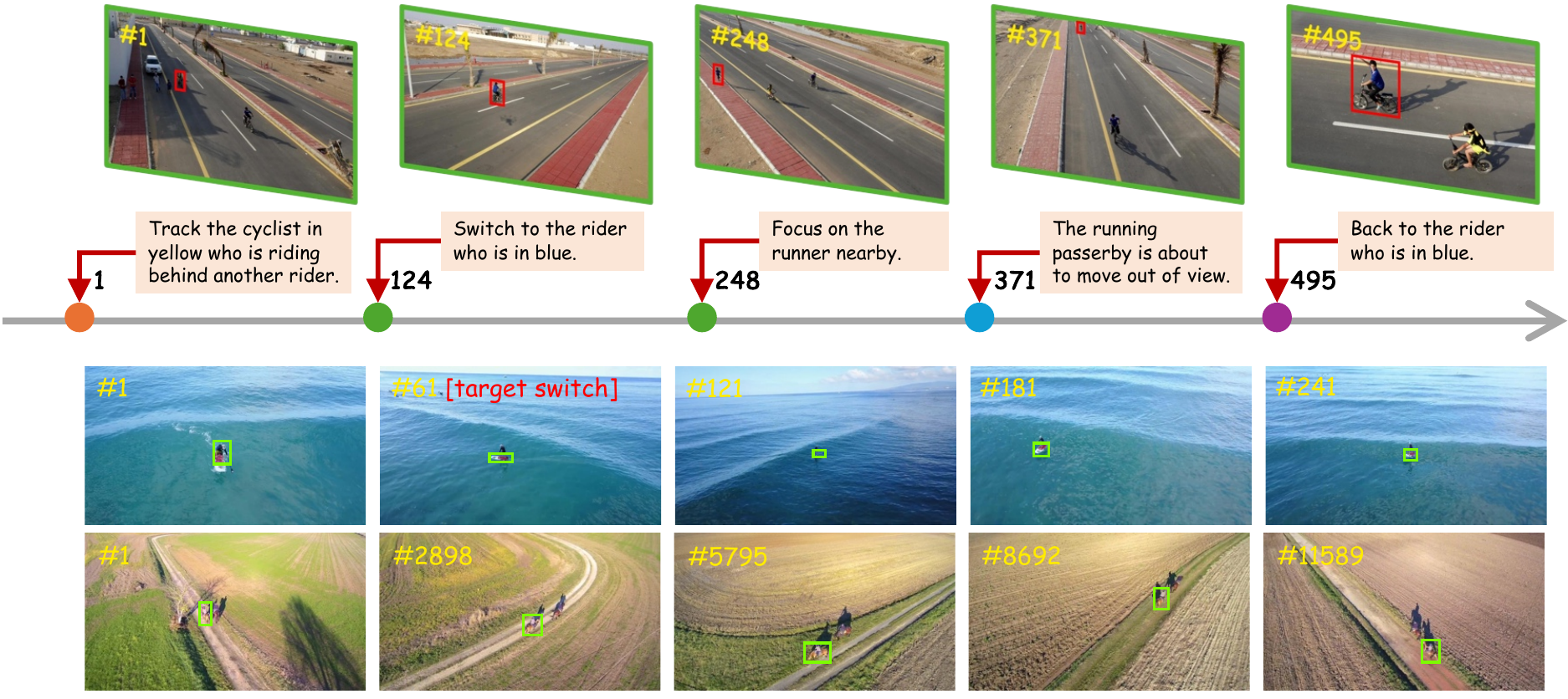}
\caption{
Representative sequences from the \textbf{UAV tracking} scenario.
This scenario includes aerial-view videos characterized by large scale variation, long-range target motion, and perspective distortions typical of drone footage.
The sequences feature vehicles, pedestrians, and small moving objects captured at varying altitudes and along diverse trajectories.
The highlighted sequence illustrates challenges such as drastic scale changes, temporary target disappearance, and viewpoint transitions during UAV maneuvers.
Other examples demonstrate additional difficulties common in aerial surveillance, including low-resolution targets, cluttered backgrounds, and complex scene geometries.
}
    \label{fig:uav}
\end{figure*}

\begin{figure*}[ht]
    \centering
    \includegraphics[width=1.0\linewidth]{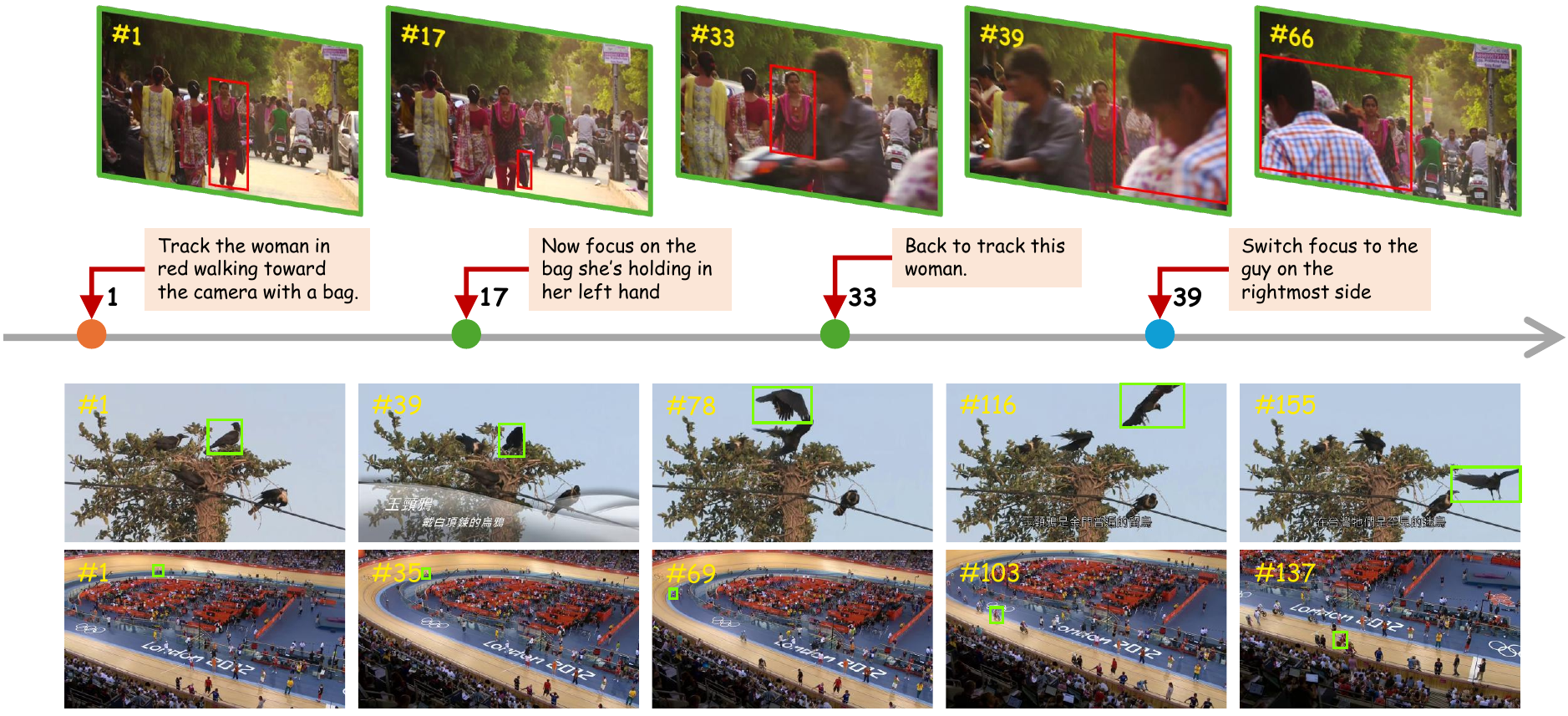}
\caption{
Representative sequences from the \textbf{surveillance} scenario.
This scenario features crowded street environments captured from a third-person viewpoint. The sequences depict interactions such as monitoring individuals in dense crowds, shifting attention between people and carried objects, and handling frequent occlusions and complex motion patterns. The highlighted sequence illustrates target-switch interaction cues.
}
    \label{fig:surveillance}
\end{figure*}

\begin{figure*}[ht]
    \centering
    \includegraphics[width=1.0\linewidth]{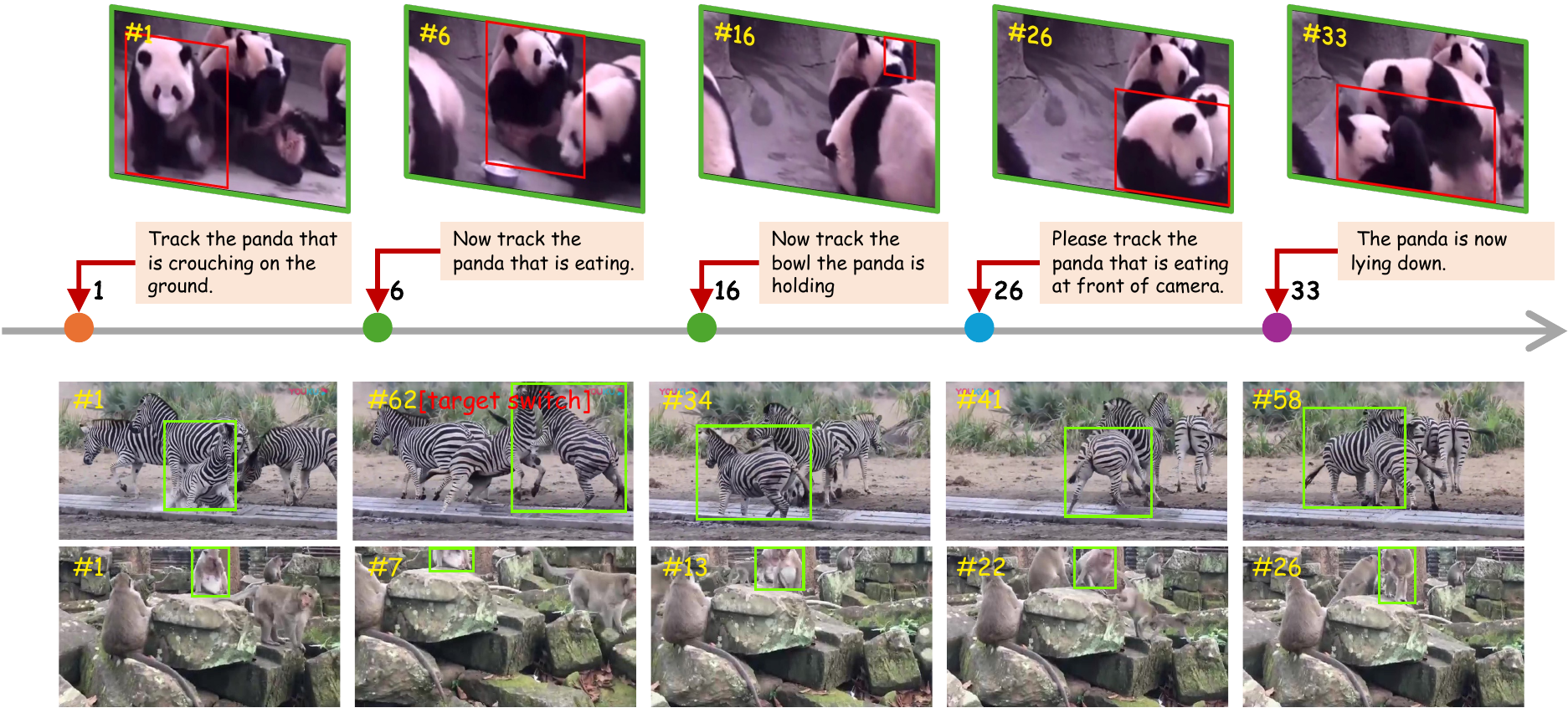}
\caption{
Representative sequences from the \textbf{wildlife monitoring} scenario.
This scenario includes animal behaviors captured in natural environments. The sequences depict interactions such as observing feeding behaviors, shifting attention between different animals, and focusing on objects being manipulated by the animals.
}
    \label{fig:wildlife_monitoring}
\end{figure*}

\begin{figure*}[ht]
    \centering
    \includegraphics[width=1.0\linewidth]{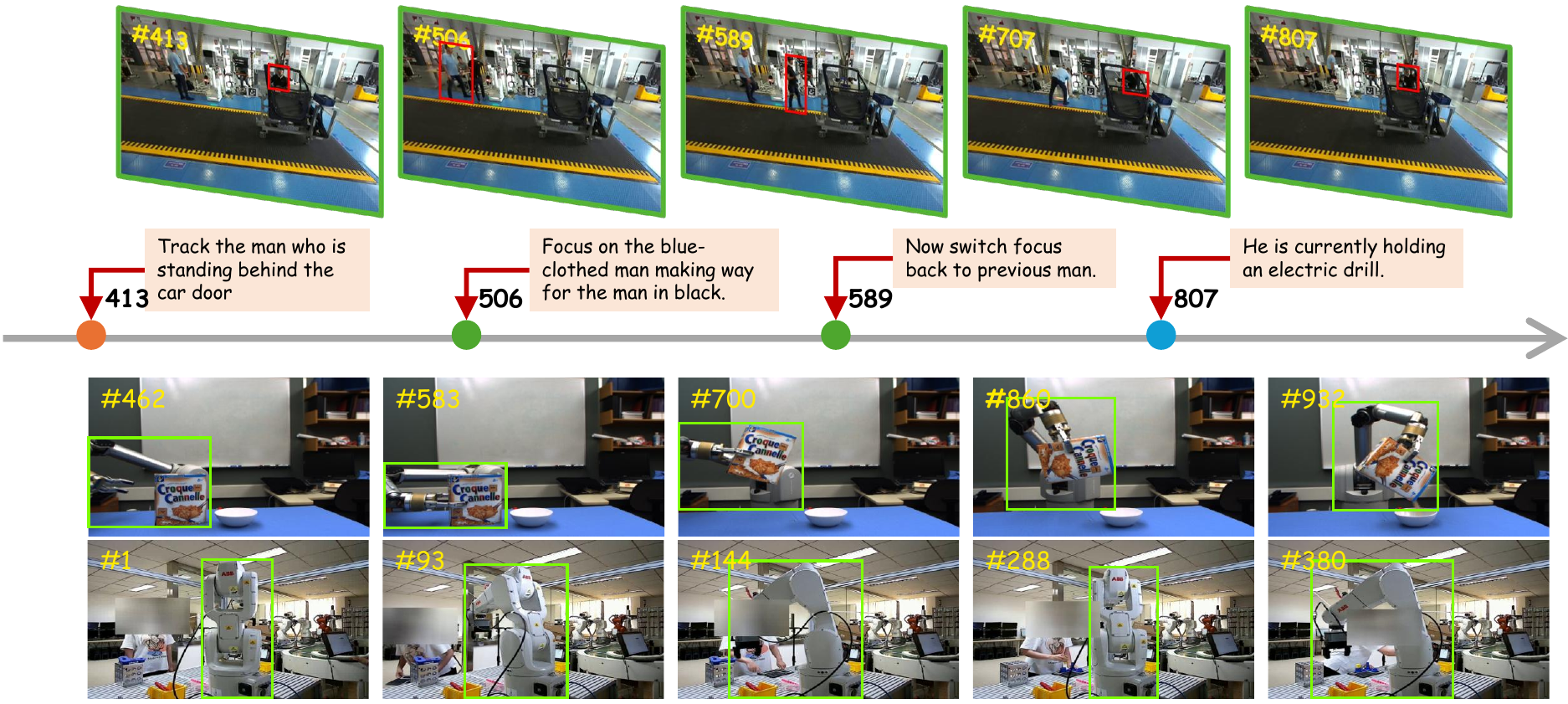}
\caption{
Representative sequences from the \textbf{other} scenario.
This scenario contains diverse scenes that do not fall into the previous categories, including industrial, laboratory, and indoor environments. The sequences illustrate various interaction points where user instructions shift attention among different people, objects, and activities.
}
    \label{fig:other}
\end{figure*}

Representative sequences from each scenario are shown in Figs.~\ref{fig:daily_activities}, \ref{fig:sports_analysis}, \ref{fig:uav}, \ref{fig:surveillance}, \ref{fig:wildlife_monitoring}, and \ref{fig:other}.

\section{Additional Experimental Results}
\begin{table}[ht]
\vspace{-4mm}
\centering
\caption{Categorization of the evaluated tracking methods by task type, architectural family, and publication venue.}
\vspace{-2mm}
\label{tab:trackers_by_category}
\renewcommand{\arraystretch}{0.8}
\begin{tabular}{llll}
\toprule
\textbf{Type} & \textbf{Tracker} & \textbf{Architecture} & \textbf{Where} \\
\midrule
\multirow{2}{*}{VLT}
  & JointNLT    & Trans. & CVPR’23 \\
  & DUTrack     & Trans. & CVPR’25 \\
  & SUTrack     & Trans. & AAAI’25 \\

\midrule

 \multirow{5}{*}{VOS} 
  & SAM2       & Trans. & ICLR'25 \\
  & HIM2SAM       & Trans. & PRCV'25 \\
  & SAMURAI       & Trans. & arxiv'25 \\
  & VL-SAM2       & Trans. & CVPRW'25 \\
  & DAM4SAM       & Trans. & CVPR'25 \\

\midrule
\multirow{17}{*}{VOT}
  & ToMP        & CNN-T  & CVPR’22 \\ 
  & SeqTrack  & Trans. & CVPR'23 \\
  & TaMOs       & Trans. & WACV'24 \\
  & SimTrack    & Trans. & ECCV’22 \\
  & CiteTracker & Trans. & ICCV’23 \\
  & STARK       & CNN-T  & ICCV’21 \\
  & MixFormer   & Trans. & CVPR’22 \\
  & OSTrack     & Trans. & ECCV’22 \\
  & DropTrack   & Trans. & CVPR’23 \\
  & ARTrack     & Trans. & CVPR’23 \\
  & GRM         & Trans. & CVPR’23 \\
  & MixViT & Trans. & PAMI’24 \\
  & ROMTrack    & Trans. & ICCV’23 \\
  & ODTrack     & Trans. & AAAI'24 \\
  & MCITrack    & Trans. & AAAI'25 \\

\midrule

  \multirow{2}{*}{RVOS}
  & VideoLisa & Trans. & NeurIPS'24 \\
  & SA2VA       & Trans. & arxiv'25 \\

\bottomrule
\vspace{-4mm}
\end{tabular}
\end{table}

\subsection{Evaluated Trackers}
Tab.~\ref{tab:trackers_by_category} summarizes the tracking methods evaluated in InteractTrack, categorized by task type (VLT, VOS, VOT, and RVOS), model architecture, and publication venue. This table highlights the breadth of contemporary tracking approaches, spanning vision–language, segmentation-based, and visual object tracking baselines. Such categorization provides a clear overview of the evaluated methods and reflects the diverse architectural designs and research directions represented in recent literature.

\subsection{Challenging Case Visualizations}

\begin{figure*}[ht]
    \centering
    \includegraphics[width=.88\linewidth]{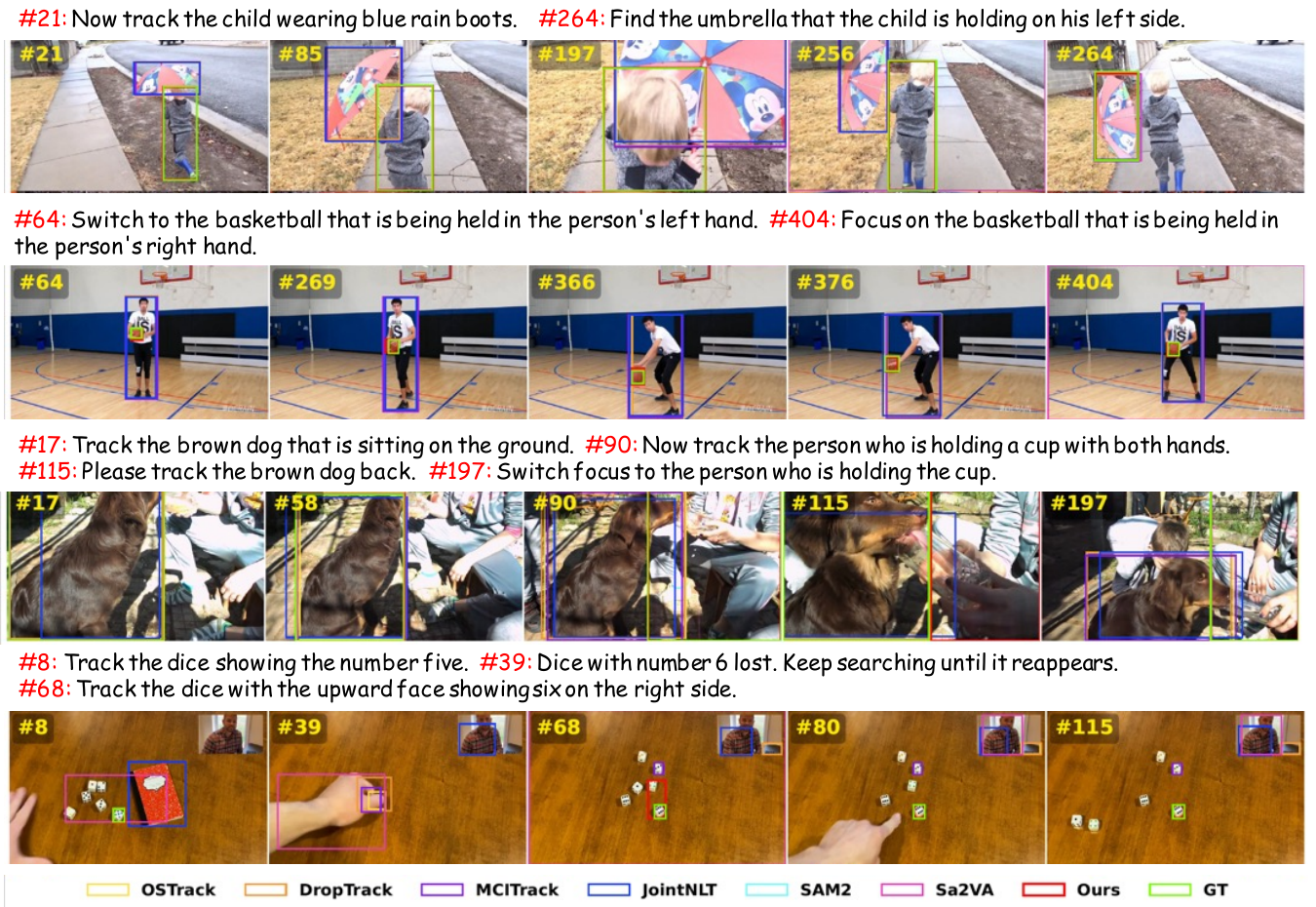}
    \caption{
\textbf{Visualization of challenging cases in InteractTrack.} It demonstrates that our proposed method can effectively handle complex interactive scenarios and accurately track the target, whereas traditional trackers may lose the target.
}
    \label{fig:vis}
\end{figure*}

Fig.~\ref{fig:vis} presents visualizations of several challenging tracking cases, including severe occlusion, small-object tracking, rapid motion, and heavy background clutter. Across these difficult scenarios, our method consistently maintains accurate localization and robust target continuity compared with existing trackers.

For example, in sequences involving occlusion or temporary target disappearance—such as the first-row examples (“the child with the umbrella” or “the dice with a missing face”)—IMAT reliably perceives and re-identifies the target once it reappears. Traditional trackers often struggle in these situations and fail to resume tracking after significant occlusion or off-screen movement.

These visual results highlight the strengths of our interaction-aware design. The IPM effectively interprets user-provided queries to ground the visual features, enabling the tracker to relocate the target even after it becomes temporarily lost. The MAVT maintains stability by adapting to appearance changes over time and filtering distractors, while the CAM ensures the tracker responds promptly to new instructions or shifts in user intent. Together, these components allow our model to handle both traditional tracking challenges and dynamic, real-time user interactions with greater robustness.

\subsection{Scenario-Based Precision}

\begin{figure*}[t]
\centering
\begin{minipage}{0.275\linewidth} 
    \centering
    \includegraphics[width=.90\linewidth]{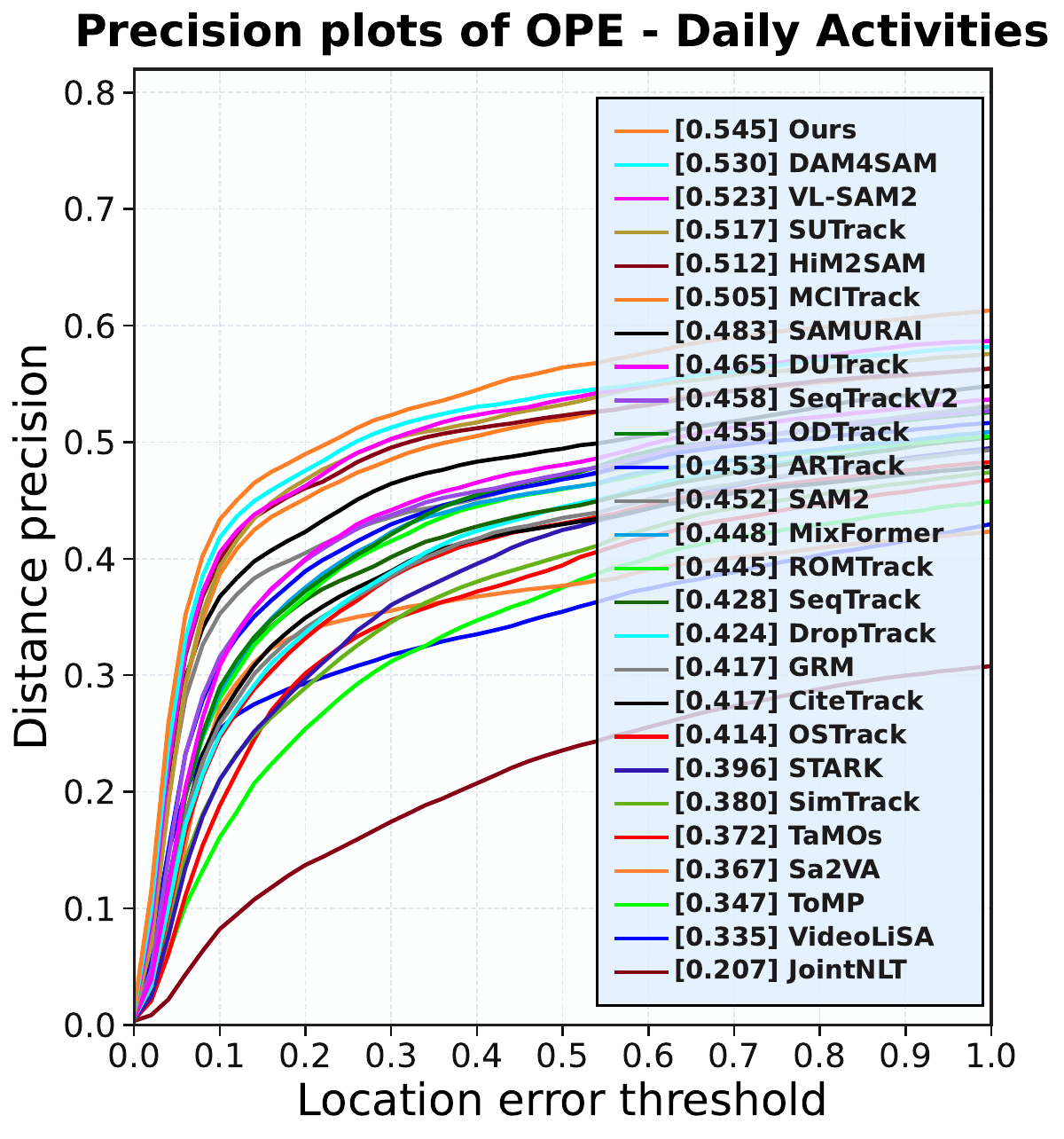}
\end{minipage}
\hspace{0.04\linewidth}
\begin{minipage}{0.28\linewidth}
    \centering
    \includegraphics[width=.90\linewidth]{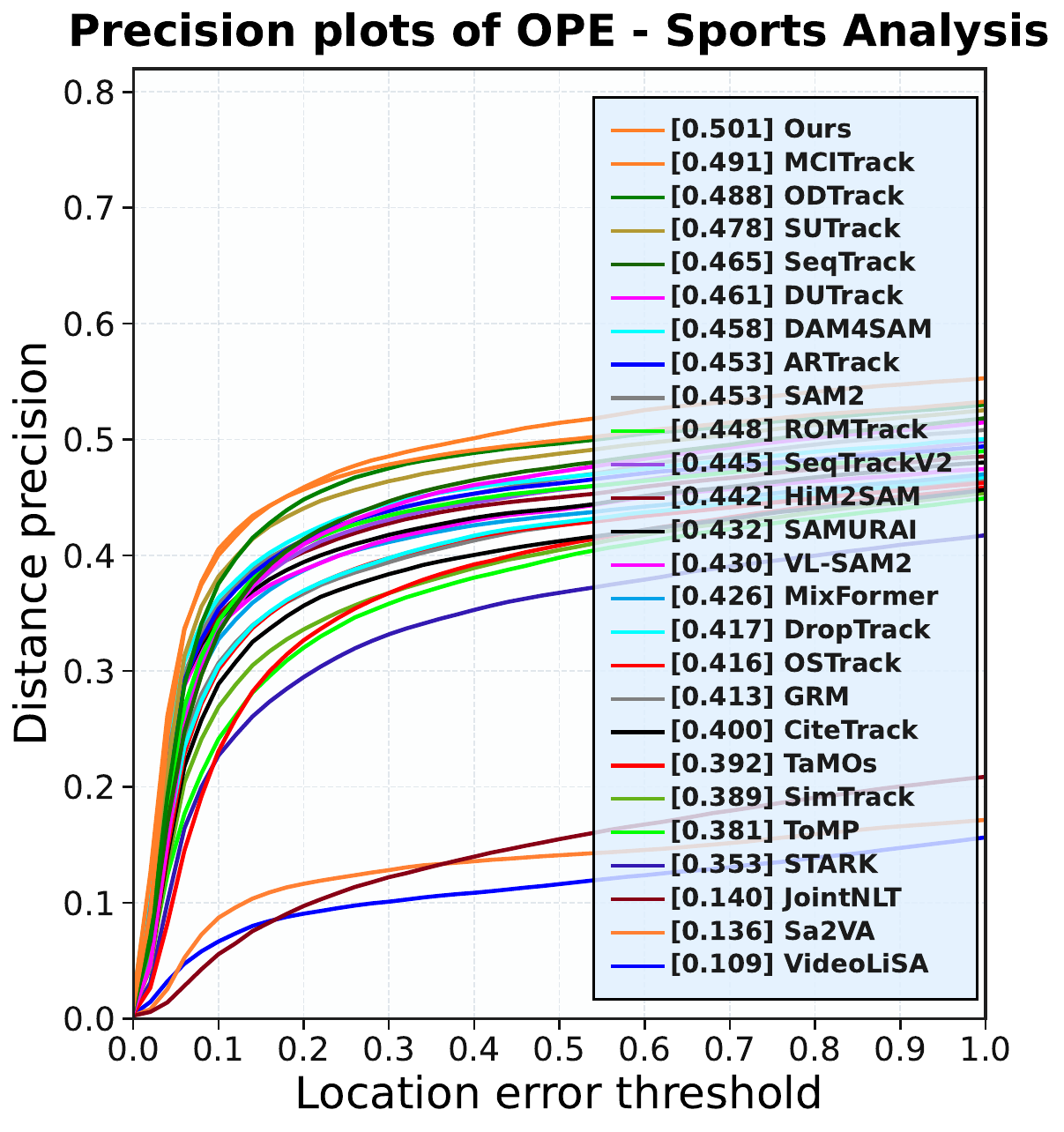}
\end{minipage}%
\hspace{0.04\linewidth}
\begin{minipage}{0.27\linewidth}
    \centering
    \includegraphics[width=.90\linewidth]{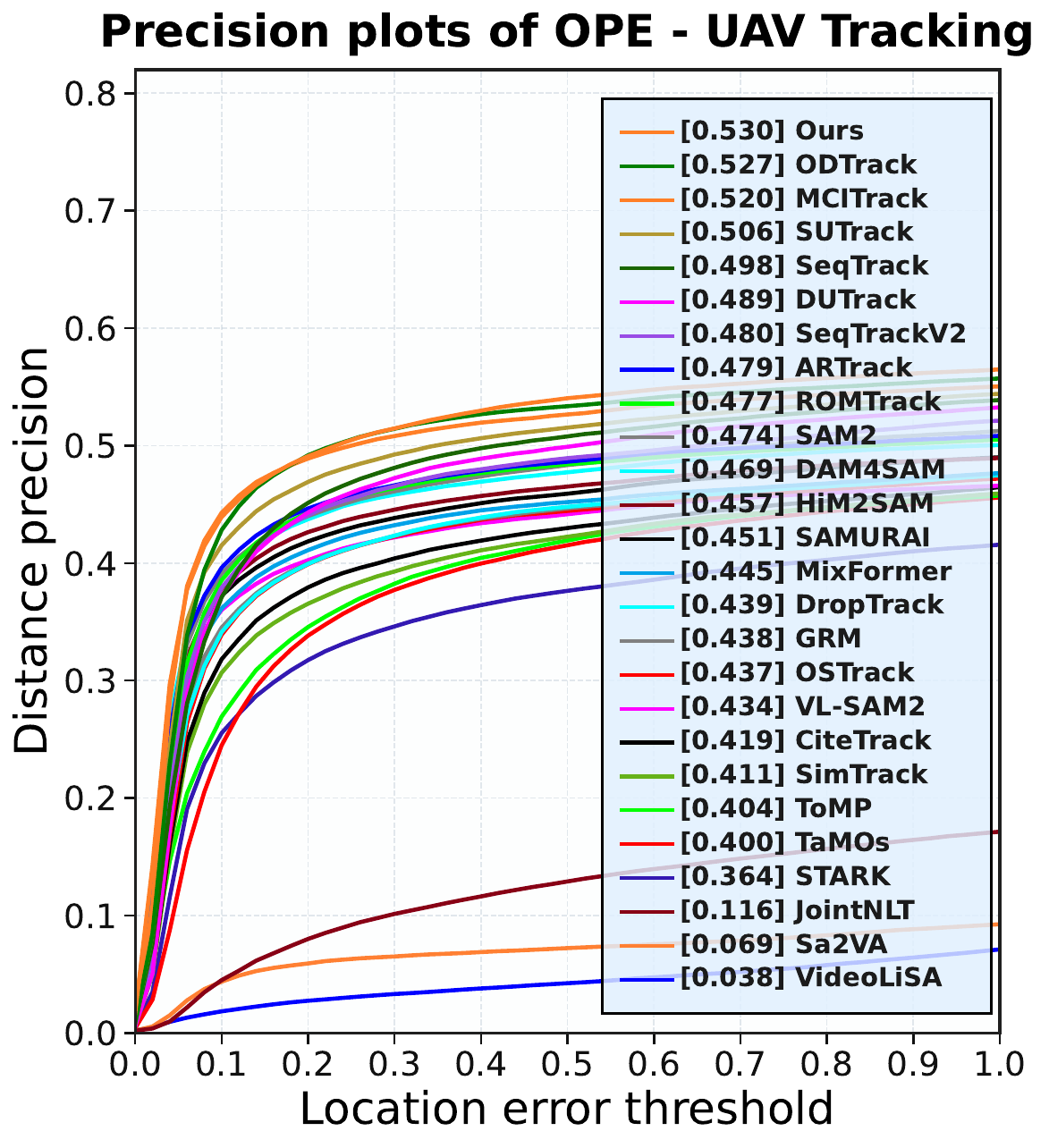}
\end{minipage}

\vspace{-1ex} 

\begin{minipage}{0.28\linewidth}
    \centering
    \includegraphics[width=.90\linewidth]{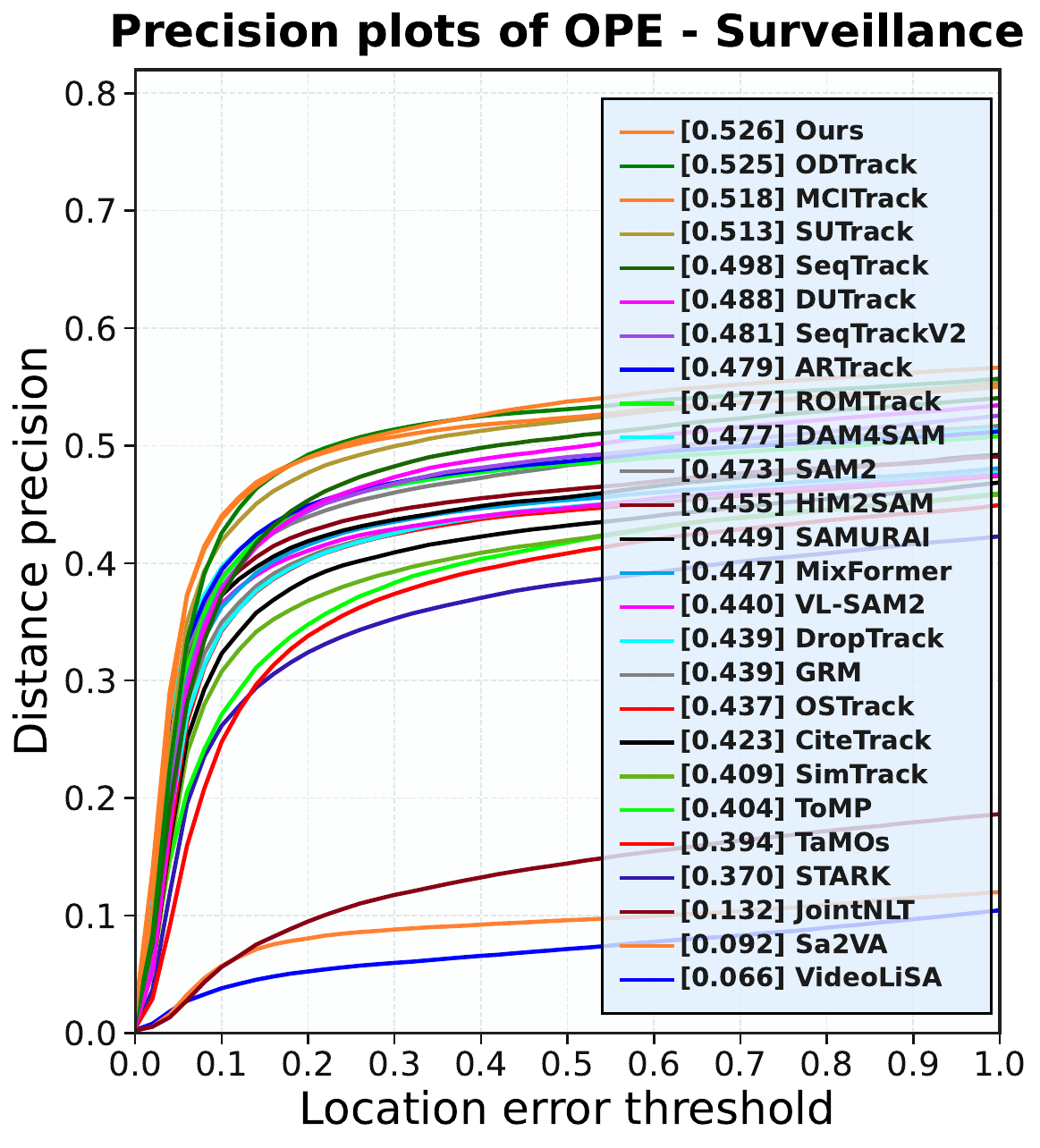}
\end{minipage}%
\hspace{0.04\linewidth}%
\begin{minipage}{0.28\linewidth}
    \centering
    \includegraphics[width=.90\linewidth]{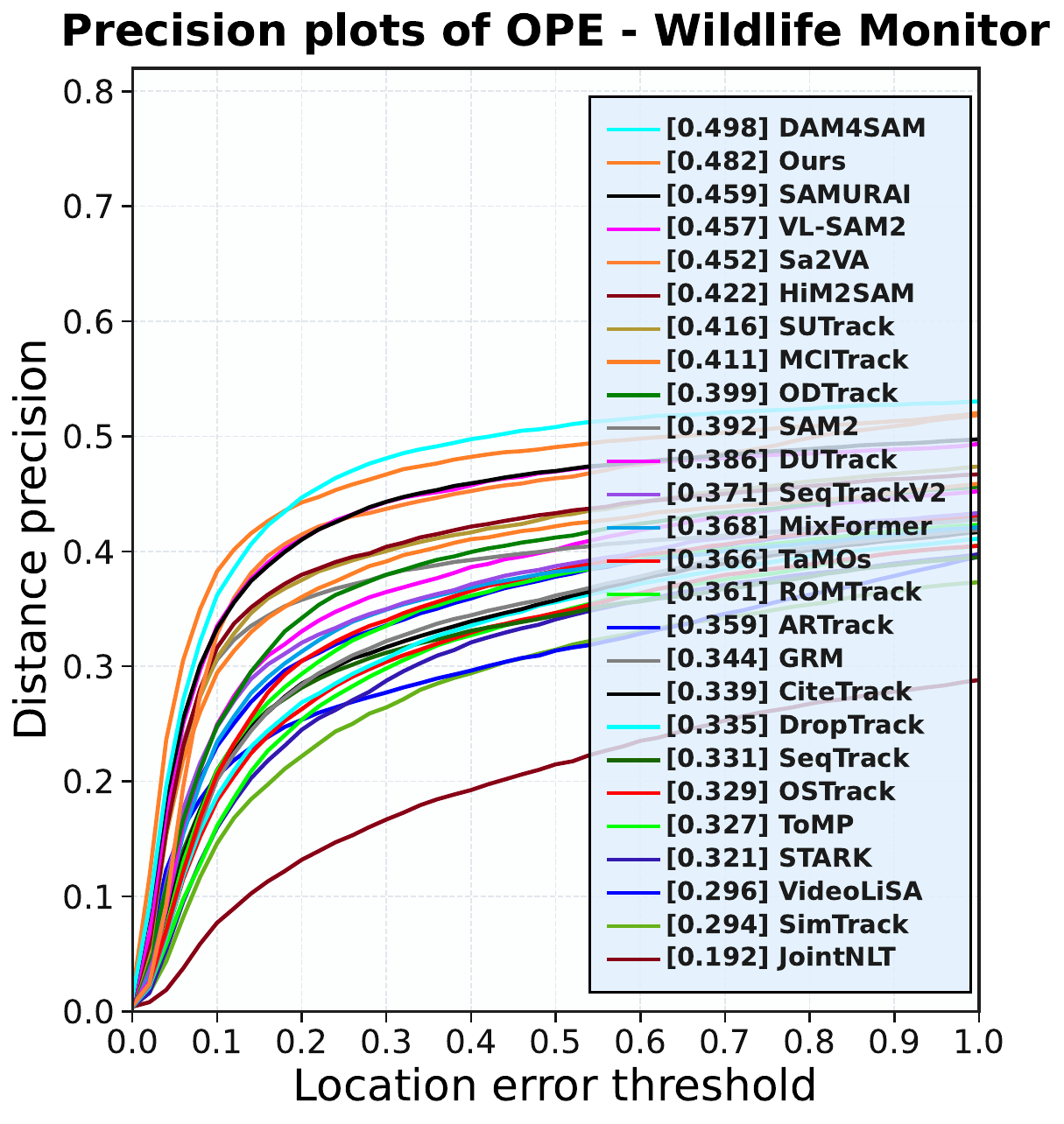}
\end{minipage}%
\hspace{0.04\linewidth}%
\begin{minipage}{0.28\linewidth}
    \centering
    \includegraphics[width=.90\linewidth]{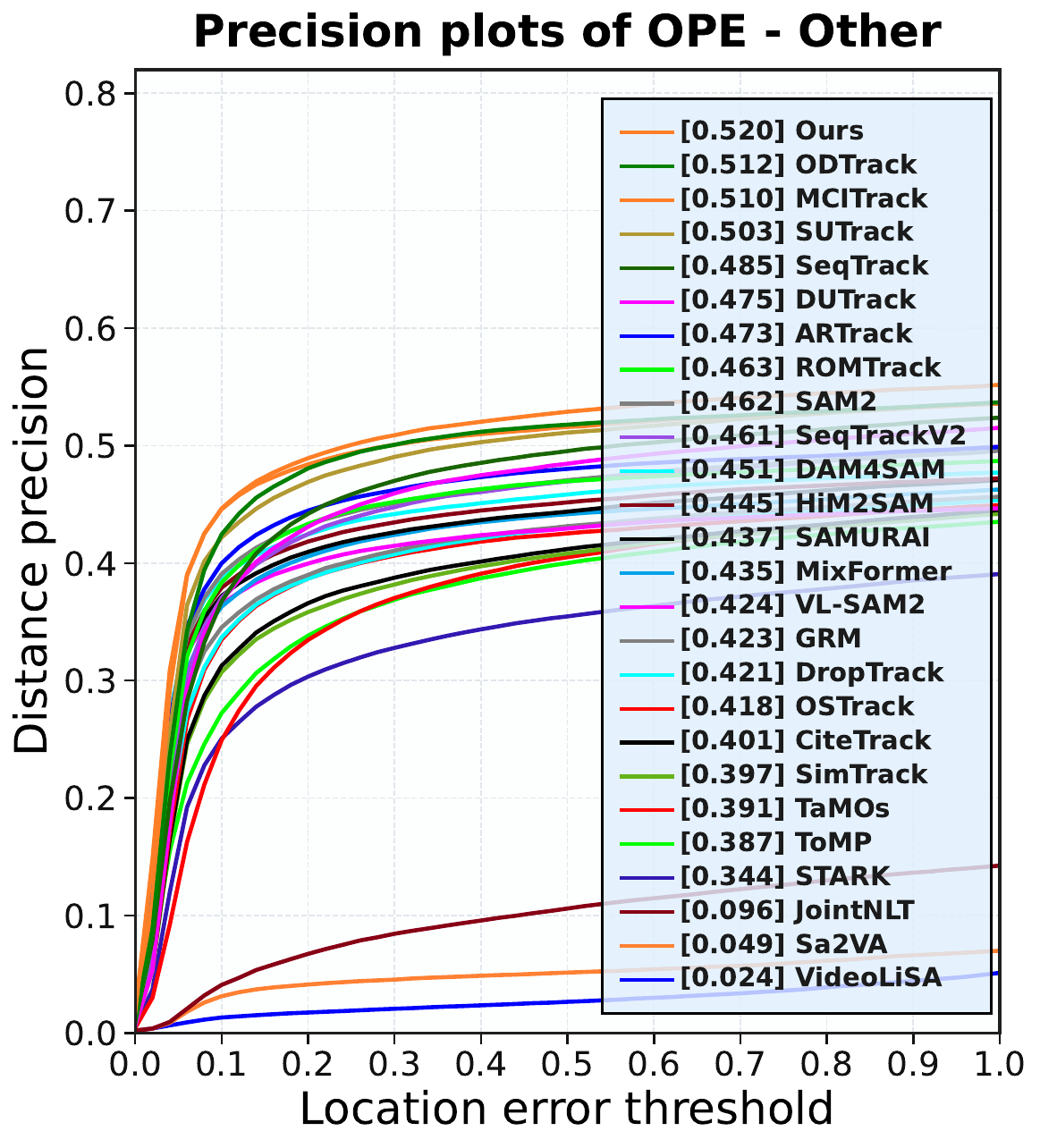}
\end{minipage}
\caption{\textbf{Precision evaluation results across the six scenarios in the InteractTrack benchmark.} The plots show that the proposed baseline achieves strong and consistent precision performance in all scenarios.}
\label{fig:scenarios-pre}
\end{figure*}

As depicted in Fig.~\ref{fig:scenarios-pre}, we further provide precision plots for all six scenarios to complement the success plots presented in the main paper. The precision evaluation offers an additional perspective on localization accuracy across varying scene conditions, revealing consistent performance trends in daily activities, sports analysis, UAV tracking, surveillance, wildlife monitoring, and other scenarios. These results reinforce the robustness and adaptability of our method across diverse environments and interaction settings.

\subsection{Failure Case Analysis}

\begin{figure*}[ht]
    \centering
    \includegraphics[width=.8\linewidth]{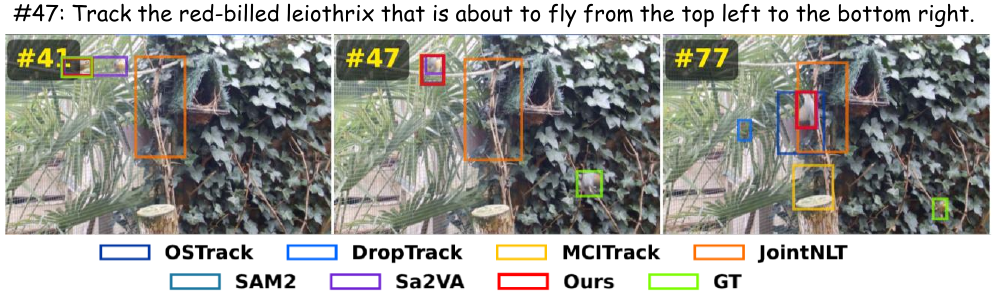}
\caption{
\textbf{Failure cases.} Even with updated user instructions, all trackers struggle with heavy occlusion and background clutter, ultimately losing the target despite interactive guidance.
}
    \label{fig:fai}
\end{figure*}

As illustrated in Fig.~\ref{fig:fai}, we present a representative failure case that highlights the inherent challenges of interactive tracking in real-world scenarios. Although updated user instructions are issued at several key frames (e.g., “Track the red-billed leiothrix that is about to fly from the top left to the bottom right”), the target undergoes rapid motion, drastic scale variation, and partial occlusions within a highly cluttered background. These factors collectively make accurate localization extremely difficult for all trackers.

This example shows that, while textual instructions can provide additional guidance, interactive tracking remains fundamentally challenging—particularly in scenarios that require the system to fuse user feedback with rapidly changing visual information. The failure case demonstrates that there is still substantial room for improvement, especially in enhancing a tracker’s adaptability to dynamic scenes, robustness to occlusion and clutter, and ability to maintain consistent contextual understanding throughout interaction-heavy sequences.

\end{document}